\documentclass[sigconf]{acmart}

\AtBeginDocument{%
  \providecommand\BibTeX{{%
    \normalfont B\kern-0.5em{\scshape i\kern-0.25em b}\kern-0.8em\TeX}}}

\setcopyright{acmcopyright}
\copyrightyear{2022}
\acmYear{2022}
\setcopyright{acmcopyright}
\acmConference[MM '22] {Proceedings of the 30th ACM International Conference on Multimedia }{October 10--14, 2022}{Lisboa, Portugal.}
\acmBooktitle{Proceedings of the 30th ACM International Conference on Multimedia (MM '22), October 10--14, 2022, Lisboa, Portugal}
\acmPrice{15.00}
\acmISBN{978-1-4503-9203-7/22/10}
\acmDOI{10.1145/3503161.3547947}

\usepackage{amssymb}
\usepackage{bbding}
\usepackage{multirow}
\usepackage{float}

\begin{document}

\title{Geometry-Aware Reference Synthesis for Multi-View Image Super-Resolution}

\author{Ri Cheng}
\orcid{0000-0002-5866-6847}
\affiliation{%
  \institution{School of Computer Science, Shanghai Key Laboratory of Intelligent Information Processing, Shanghai Collaborative Innovation Center of Intelligent Visual Computing, Fudan University}
  \city{Shanghai}
  \country{China}
}
\email{rcheng20@fudan.edu.cn}

\author{Yuqi Sun}
\orcid{0000-0002-7179-5045}
\affiliation{%
  \institution{School of Computer Science, Shanghai Key Laboratory of Intelligent Information Processing, Shanghai Collaborative Innovation Center of Intelligent Visual Computing, Fudan University}
  \city{Shanghai}
  \country{China}
}
\email{yqsun20@fudan.edu.cn}

\author{Bo Yan}
\orcid{0000-0003-0256-9682}
\authornote{Corresponding Author. This work is supported by NSFC (Grant No.: U2001209, 61902076) and Natural Science Foundation of Shanghai (21ZR1406600).}
\affiliation{%
  \institution{School of Computer Science, Shanghai Key Laboratory of Intelligent Information Processing, Shanghai Collaborative Innovation Center of Intelligent Visual Computing, Fudan University}
  \city{Shanghai}
  \country{China}
}
\email{byan@fudan.edu.cn}

\author{Weimin Tan}
\orcid{0000-0001-7677-4772}
\affiliation{%
  \institution{School of Computer Science, Shanghai Key Laboratory of Intelligent Information Processing, Shanghai Collaborative Innovation Center of Intelligent Visual Computing, Fudan University}
  \city{Shanghai}
  \country{China}
}
\email{wmtan@fudan.edu.cn}

\author{Chenxi Ma}
\orcid{0000-0002-5577-5773}
\affiliation{%
  \institution{School of Computer Science, Shanghai Key Laboratory of Intelligent Information Processing, Shanghai Collaborative Innovation Center of Intelligent Visual Computing, Fudan University}
  \city{Shanghai}
  \country{China}
}
\email{17210240039@fudan.edu.cn}

\renewcommand{\shortauthors}{Ri Cheng and Yuqi Sun, et al. }



\begin{abstract}

    Recent multi-view multimedia applications struggle between high-resolution (HR) visual experience and storage or bandwidth constraints. Therefore, this paper proposes a Multi-View Image Super-Resolution (MVISR) task. It aims to increase the resolution of multi-view images captured from the same scene. One solution is to apply image or video super-resolution (SR) methods to reconstruct HR results from the low-resolution (LR) input view. However, these methods cannot handle large-angle transformations between views and leverage information in all multi-view images. To address these problems, we propose the MVSRnet, which uses geometry information to extract sharp details from all LR multi-view to support the SR of the LR input view. Specifically, the proposed Geometry-Aware Reference Synthesis module in MVSRnet uses geometry information and all multi-view LR images to synthesize pixel-aligned HR reference images. Then, the proposed Dynamic High-Frequency Search network fully exploits the high-frequency textural details in reference images for SR. Extensive experiments on several benchmarks show that our method significantly improves over the state-of-the-art approaches.

    
\end{abstract}





\maketitle

    \begin{figure}[t]
    \centering
    \includegraphics[width=\linewidth]{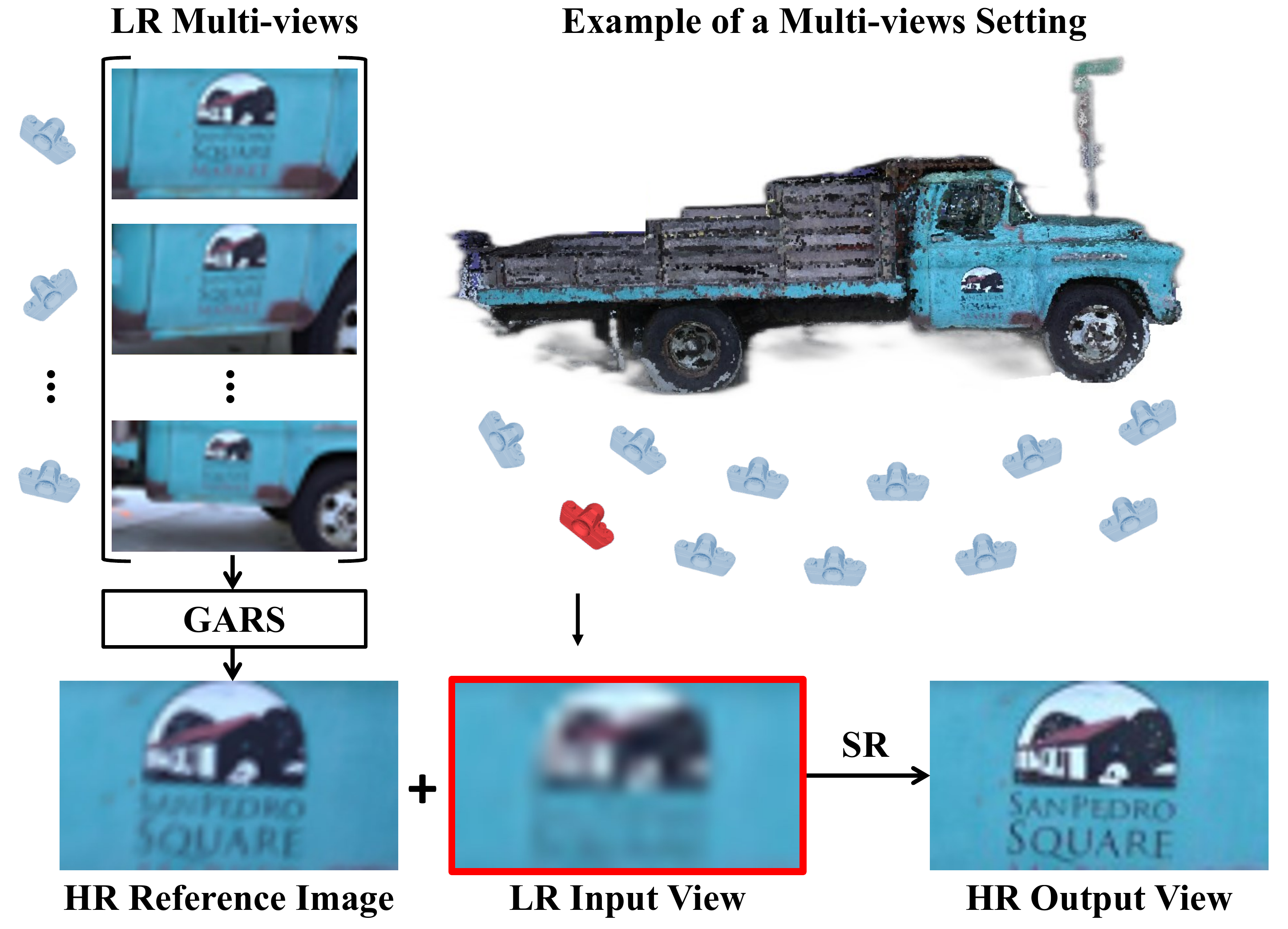}
    \caption{\textbf{Multi-view image super-resolution.}
  We propose GARS module for extracting the HR details from all \emph{LR Multi-View}. Then, our GARS module synthesizes high-frequency textural details into \emph{HR Reference Image} that is aligned with the bicubic of \emph{LR Input View}. The word SQUARE is more textured in \emph{HR Reference Image} than in the \emph{LR Input View}.
    }
    \label{001_intro_image_mm}
    \end{figure}

\section{Introduction}
\label{sec:intro}

    
    Multi-view Multimedia (MVMM) uses numerous cameras to capture multi-view images of the same scene from various angles. 
    It is widely used in auto-driving, street-view navigation, free-viewpoint video\cite{4037061, lee_tabatabai_tashiro_2015}, and virtual reality\cite{799723, article_vr2}.
    Recent MVMM applications call for high-resolution (HR) images to give a pleasing visual experience and assist downstream tasks such as detection and tracking.
    Nevertheless, HR multi-view images face substantial storage costs or bandwidth constraints.  
    To address this problem, we propose the Multi-View Image Super-Resolution (MVISR) task, which increases the resolution of multi-view images by using all low-resolution (LR) multi-view images in the same scene.
    
    
    A naive approach to MVISR is to perform single image SR (SISR) \cite{7115171,Lim_2017_CVPR_Workshops} on each view. They take a single view as input and cannot utilize the relationship across all LR views. Other alternatives include video SR (VSR) methods \cite{10.1007/978-3-030-58607-2_20,Chan_2021_CVPR}. They regard multi-views as video frames with little object motion, and so cannot deal with the large-angle transformations between input views.  Some reference-based SR (Ref-SR) methods \cite{Zhang_2019_CVPR,Yang_2020_CVPR,Lu_2021_CVPR} can overcome angle constraints and extend the range of reference views. However, Ref-SR is ineffective due to the difficulties involved with reference image selection in all LR multi-views and reference alignment when the scale and rotation differences between the Ref and LR input are large. In the end, none of the approaches discussed previously can explicitly leverage all LR multi-views and their associated geometry information.

    Therefore, in this paper, we propose MVSRnet, which utilizes the geometry information to exploit high-frequency (HF) reference information from all LR views for each view SR. A camera that is closer to the objects can capture sharper images than a distant one. 
    It inspires us that a close-up view contains more textural details that can provide sufficient reference for the distant view. To realize this idea, we design Geometry-Aware Reference Synthesis (GARS) module in MVSRnet. As illustrated in Figure~\ref{001_intro_image_mm}, the word SQUARE is blurred in the LR input view but sharp in other LR views. Our GARS module takes LR multi-views as input to generate reference images with rich textural details and well-aligned with the LR input view. To be specific, we use structure-from-motion \cite{Schonberger_2016_CVPR} and multi-view stereo \cite{10.1007/978-3-319-46487-9_31,Yao_2018_ECCV} to extract scene geometry information such as the camera pose and depth map. After that, GARS conducts 3D image warping to align all LR multi-views into the target view. 
    Then, we introduce a depth-guided patch-selection strategy to select close views on the patch level and fuse them into HR multi-view reference images (MVRs).

    \begin{figure}[t]
    \centering
    \includegraphics[width=\linewidth]{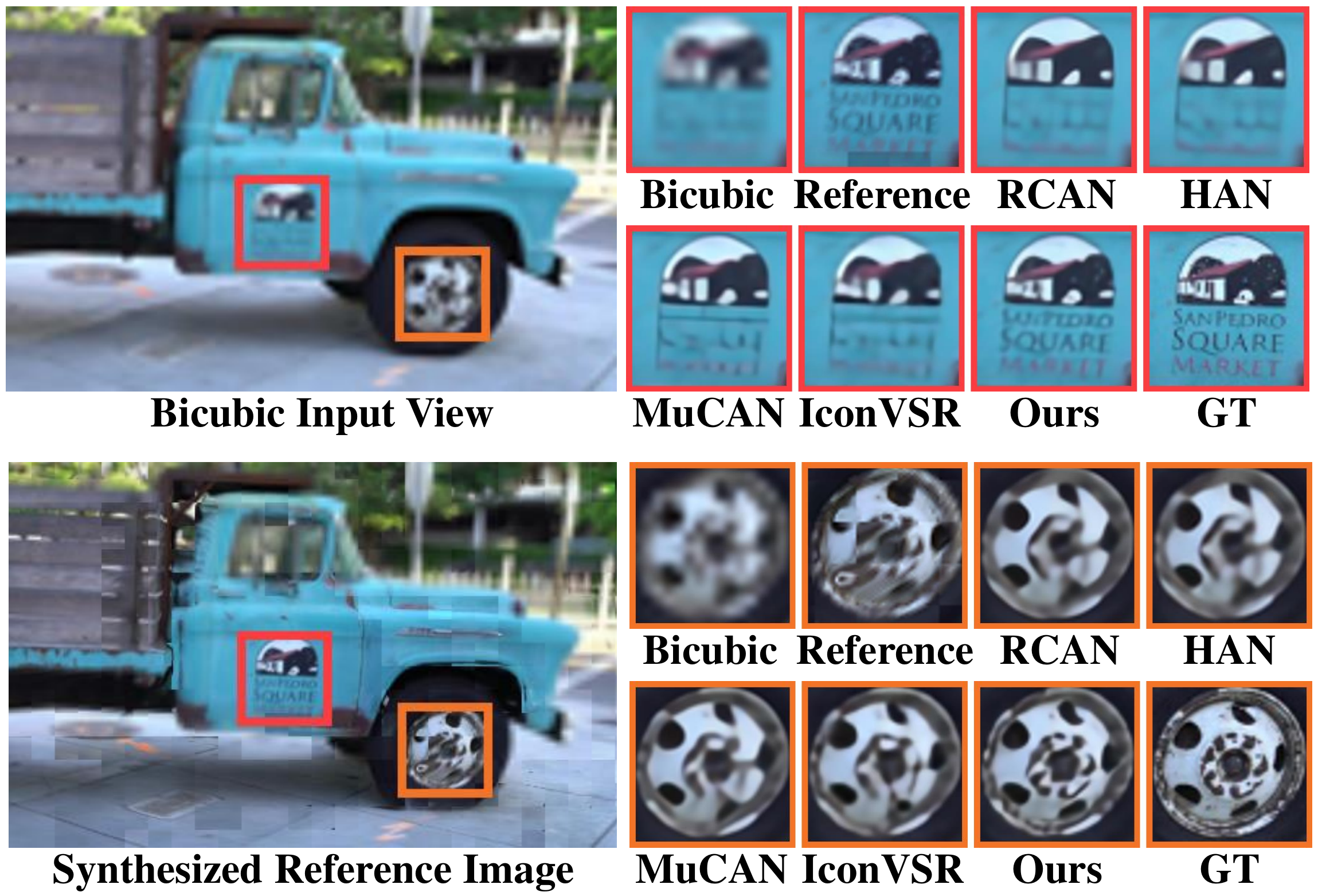}
    \caption{\textbf{Example of multi-view image super-resolution results.}
    Our proposed model outperforms SISR (RCAN \cite{zhang2018rcan} and HAN \cite{10.1007/978-3-030-58610-2_12}) and VSR (MuCAN \cite{10.1007/978-3-030-58607-2_20} and IconVSR \cite{Chan_2021_CVPR}) methods in visual quality. 
    For instance, by utilizing the synthesized reference image, our model is able to recover finer details of the word SQUARE in the first example and the center area of the tire in the second example.
    }
    \label{002_intro_experiment_mm}
    \end{figure}
    
    On the other hand, MVRs synthesized from all LR views in GARS have artifacts and some pixels that are not properly aligned with the LR input view due to inaccuracies in the geometry information. 
    Nearby views have a small difference in viewing angle and hence suffer less from geometric inaccuracies.
    As a result, we also synthesize near-view reference images (NVRs) with fewer artifacts but fewer HF textural details using nearby views. Then, in order to properly leverage MVRs and NVRs, we design the Raw Selection Module (RSM) and Adaptation Selection Module (ASM) in our Dynamic High-Frequency Search (DHFS) network to exploit the HF textural details by reducing the feature difference dynamically. Finally, we perform experiments to demonstrate the proposed MVSRnet SR capability, and some comparative experiments are shown in Figure~\ref{002_intro_experiment_mm}. 
    Additionally, our work contributes to the Ref-SR methodology by proposing a way for performing SR using multi-view LR images rather than HR images  \cite{Zhang_2019_CVPR} as reference.

    This paper mainly has the following contributions:
    \begin{itemize}
    \item For the first time, this paper proposes the MVSRnet, a learning-based network that makes explicit use of geometric information to address the MVISR problem. 
    In addition, this work enriches the Reference-Based SR research.

    \item The Geometry-Aware Reference Synthesis module synthesizes the HR reference image from all LR multi-view images using geometry information. 
    The reference supports SR by containing rich high-frequency textural details and aligning with the input view.

    \item The Dynamic High-Frequency Search network is proposed for reconstructing the high-resolution view by dynamically
    reducing the feature difference to exploit high-frequency textural details from synthesized reference images.
    
    
    
    \end{itemize}

\section{Related Work}
\label{sec:related_work}

\subsection{Single Image Super-Resolution}


Single Image Super-Resolution (SISR) \cite{7115171,Lim_2017_CVPR_Workshops, zhang2018rcan,Ledig_2017_CVPR,Wang_2018_ECCV_Workshops,Ma_2020_CVPR} aims to learn the mapping function between LR and HR images. 
However, SISR is unable to recover HF details due to the destruction of HR textural clues in the downsampling LR image.

    \begin{figure*}[t]
    \centering
    \includegraphics[width=0.95\linewidth]{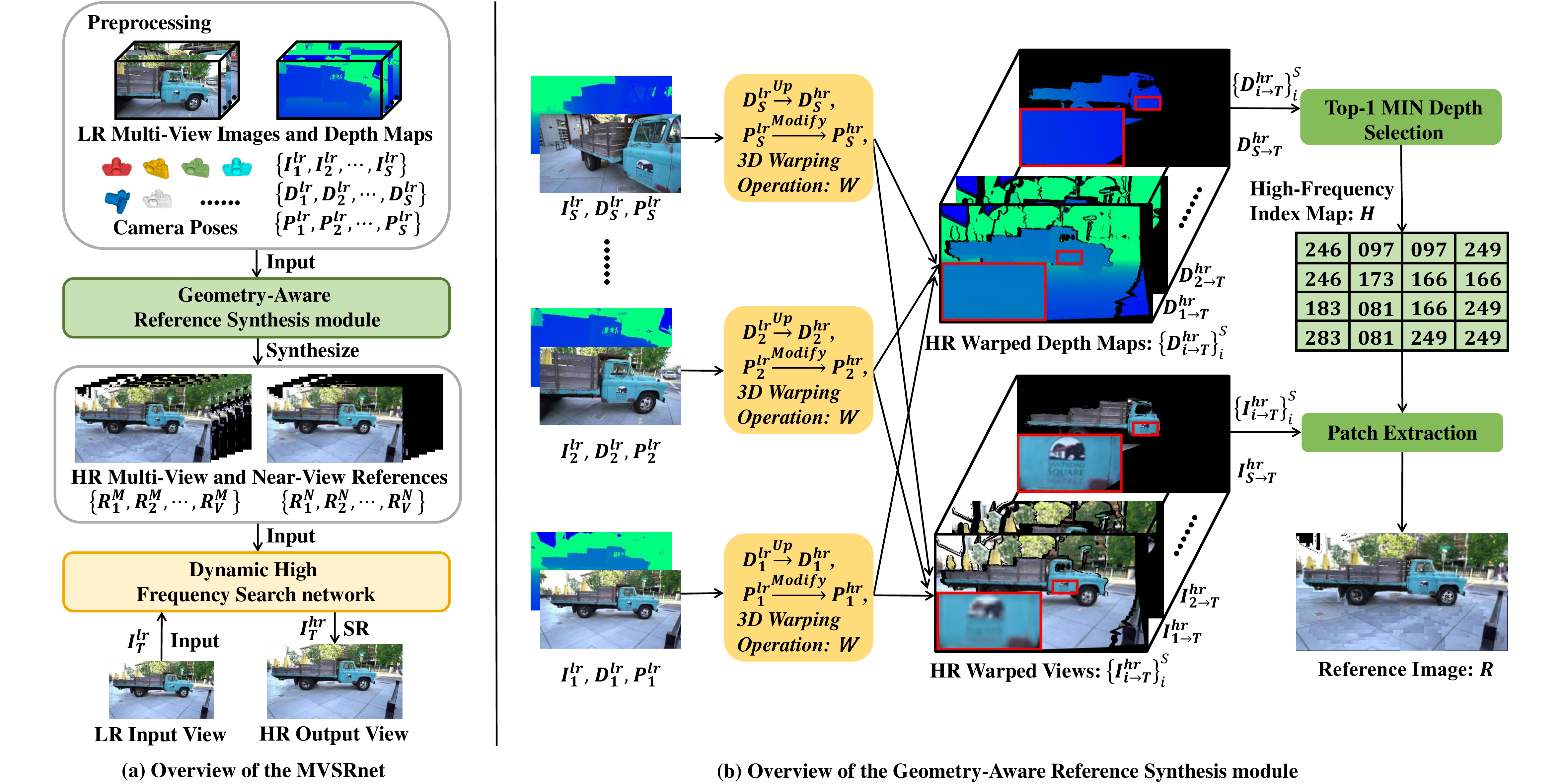}
    \caption{
    (a) The overview pipeline of the proposed MVSRnet.
    (b) The depth value of truck door area in the warped depth map $D_{S \rightarrow T}^{h r}$ is smaller (darker blue) than $D_{1 \rightarrow T}^{h r}$, so the warped view $I_{S \rightarrow T}^{h r}$ contains more  truck door HR details than $I_{1 \rightarrow T}^{h r}$, corresponding to more truck door high-frequency details in view $I_S^{l r}$ than $I_1^{l r}$.
    }
    \label{003_004_overview_GARS}
    \end{figure*}

\subsection{Reference-Based Super-Resolution}

Reference-based super-resolution (Ref-SR) approaches transfer the HR details of the reference image to the LR input image. 
Several works \cite{Shim_2020_CVPR,Zheng_2018_ECCV} leverage deformable convolution \cite{Dai_2017_ICCV,Zhu_2019_CVPR} or estimated flow to align and extract reference features. 
However, it will seriously affect the performance of these algorithms when finding long-distance correspondence. 
The patch-based Ref-SR approaches \cite{Zhang_2019_CVPR,10.1007/978-3-030-58595-2_4,Yang_2020_CVPR,Lu_2021_CVPR,Jiang_2021_CVPR} perform patch matching in the feature space by calculating the cosine similarity between the LR patch and Ref patch.  
\cite{Yang_2020_CVPR, Lu_2021_CVPR} formulate patch matching based on the transformer \cite{NIPS2017_3f5ee243}.
However, the difference in scale and rotation between the HR Ref and LR inputs can cause matching errors \cite{Jiang_2021_CVPR}, which is serious in multi-view settings. 
In addition, for the super-resolution of multi-view LR images, there is no available view reference selection method. 
The SIFT-based \cite{790410} selection method used in constructing the CUFED5 \cite{Zhang_2019_CVPR} dataset is only applicable to selecting HR references whose content is similar to the LR input view.

\subsection{Multi-View Super-Resolution}

Several traditional approaches \cite{5459378,Tsiminaki_2014_CVPR} and learning-based approaches \cite{Li_2019_CVPR,8885943} study the multi-view texture SR problem. 
However, they only super-resolve the texture of the object instead of the entire image.
Stereo image SR \cite{8998204, Wang_2019_CVPR_stereo} utilizes additional information provided from a second viewpoint, but it uses a much smaller number of viewpoint than MVISR.
Light field image SR  \cite{5559010,6574844,Zhang_2019_CVPR_2,10.1007/978-3-030-58592-1_18} has also been extensively studied in recent years, but the view span are limited compared to the unstructured multi-views.
Our MVSRnet differs from multi-view video SR \cite{9286862, 10.1145/2964284.2967260} in that they utilize HR reference frames, whereas MVSRnet exclusively uses LR frames. However, obtaining HR images is difficult due to substantial storage costs and bandwidth constraints in reality.
Sun \emph{et al.} \cite{DBLP:conf/mm/SunC0Z21} addresses the space-angle SR problem, but only two adjacent views are used to synthesize virtual view and SR.
In general, it is difficult to apply the methods discussed above to the MVISR task directly.

\section{Proposed Method}
\label{sec:our_approach}

    This section describes the proposed network called MVSRnet, and the overview of the MVSRnet is shown in Figure~\ref{003_004_overview_GARS}(a).
    We begin with a data preprocessing step that estimates depth maps $\{D_i^{l r}\}_{i=1}^S$ and poses $\{P_i^{l r}\}_{i=1}^S$ of all $S$ LR multi-views in Section~\ref{sec:Preprocessing}.
    Then, in Section~\ref{sec:GARS}, we introduce Geometry-Aware Reference Synthesis (GARS) that uses geometry priors ($\{D_i^{l r}\}_{i=1}^S$ and $\{P_i^{l r}\}_{i=1}^S$) to synthesize $V$ HR multi-view reference (MVR) $\{R_i^{M}\}_{i=1}^V$ and HR near-view reference (NVR) $\{R_i^{N}\}_{i=1}^V$.
    MVR stores more high-frequency (HF) textural details than NVR since MVR is synthesised from all $S$ LR multi-view images $\{I_i^{l r}\}_{i=1}^S$, not from $L$ nearby views like NVR.
    However, there are fewer artifacts caused by erroneous warping in the NVR.
    Therefore, in Section~\ref{sec:DHFS}, we design Dynamic High-Frequency Search (DHFS) network that dynamically selects and extracts the correct HF details in MVRs and NVRs to super-resolve the LR input view $I_T^{l r}$ and get HR output view $I_T^{h r}$. 
    At last, loss functions are shown in Section~\ref{sec:loss}.

\subsection{Preprocessing}
\label{sec:Preprocessing}
    
    To obtain the geometry information of the multi-view scene, we utilize COLMAP \cite{Schonberger_2016_CVPR} to compute camera poses by inputting all $S$ LR multi-views. This is known as structure-from-motion (SfM), and we can get the camera pose $\{P_i^{l r}\}_{i=1}^S$ of each viewpoint. The pose is represented by the extrinsic parameters (rotation matrices and translation vectors) and intrinsic parameters. 
    We use the depth map of the LR input view to align all LR multi-views to the LR input view. Therefore, we conduct the multi-view stereo (MVS) in COLMAP to obtain the depth map for each LR view $\{D_i^{l r}\}_{i=1}^S$ by inputting all LR multi-views and the SfM output.

\subsection{Geometry-Aware Reference Synthesis}
\label{sec:GARS}

    The GARS module synthesizes references (Refs) from all multi-view images, and we display the synthesis process of a Ref in         Figure~\ref{003_004_overview_GARS}(b).
    The reference image is aligned with the input view and contains high-frequency details.
    
    \noindent
    \textbf{Alignment Using Scene Geometry.} When the reference image has a large deformation in comparison to the input image, the alignment capabilities of recent Ref-SR approaches are reduced.
    For instance, Figure~\ref{003_004_overview_GARS} illustrates a great differences in scale and rotation of the truck between the LR input view $I_T^{l r}$ and view $I_S^{l r}$.
    Therefore, we use the multi-view scene geometry to explicitly align all multi-view source images $\{I_i^{l r}\}_{i=1}^S$ with the input view $I_T^{l r}$.
    Specifically, we up-sample the depth map to the target output resolution (SR factor is $\times4$ in our experiment setting) based on the bicubic mode and modify the camera intrinsic matrices according to the up-sampling factor.
    Then we apply the 3D warping operation \cite{mcmillan1997image} to obtain warped views $\{I_{i \rightarrow T}^{h r}\}_{i=1}^S$ as follows: 
    \begin{equation}
    I_{i \rightarrow T}^{h r}=W\left(D_{i}^{h r}, D_{T}^{h r}, P_{i}^{h r}, P_{T}^{h r}, I_i^{l r}\right),
    \label{equ:000_3D_warping_operation}
    \end{equation}
    where $i$ and $T$ refer to $i$-th multi-view images and input view. 
    $D^{h r}$ and $P^{h r}$ denote the depth maps and the camera pose at the same resolution as the target output solution.
    Specifically, the pixel coordinate correspondence between the multi-view images and the input view in the 3D warping operation is as the following formula:
    \begin{equation}
    D_i(p_i) * p_i = K_i ( R_r K_T^{-1} D_T(p_T) * p_T + t_r),
    \label{equ:001_pixel_coordinate_correspondence}
    \end{equation}
    where $p$ denotes pixel homogeneous coordinates.
    Camera intrinsic matrices $K$, rotation matrices $R$ and translation matrices $t$ are provided by camera posed $P$ as presented in Eq.(\ref{equ:000_3D_warping_operation}). $R_r = R_i R_T^T$, $t_r = t_i - R_r t_T$ are relative rotation and translation matrices.
    
    It is noted that we can directly apply 3D warping operation on LR depth maps without up-sampling and obtain LR aligned views $\{I_{i \rightarrow T}^{lr}\}_{i=1}^S$. 
    However, the general sampling modes in warping operation, such as nearest and bilinear, blurs and destructs high-frequency details using the weighted average method when dealing with overlapping pixels.
    One possible solution is to use a pre-trained network as a feature extractor to conduct 3D warping operations in the LR feature space.
    However, this leads to both spatially and temporally inefficient as hundreds of views need to be extracted features in each training iteration.
    
    \noindent
    \textbf{HF Details Extraction Using Depth-Guided Patch-Selection Strategy.}
    We should extract useful HF details from warped views $\{I_{i \rightarrow T}^{h r}\}_{i=1}^S$ that contain redundant details as the warped view contains the LR details area as well.
    We assume that the view taken at close range has more high-frequency textural details than the distant view.
    Therefore, we apply the 3D warping operation on source depth maps $W(D_{i}^{h r}, D_{T}^{h r}, P_{i}^{h r}, $ $P_{T}^{h r}, D_i^{l r})$ to obtain HR warped depth maps $\{D_{i \rightarrow T}^{h r}\}_{i=1}^S$. 
    Warped depth maps present distance geometry information, i.e., the distance between the object and the camera.
    The smaller the depth value, the closer the object in source view $I_i^{l r}$ is to the camera. As a result, we can leverage the HR warped depth maps to locate the HR details area in the warped views $\{I_{i \rightarrow T}^{h r}\}_{i=1}^S$. 
    For example, Figure~\ref{003_004_overview_GARS}(b) shows that the truck door has smaller depth values (darker blue) in the warped depth map $D_{S \rightarrow T}^{h r}$ than $D_{1 \rightarrow T}^{h r}$. 
    Hence, the HR warped view $I_{S \rightarrow T}^{h r}$ contains more truck door HR details than $I_{1 \rightarrow T}^{h r}$, corresponding to more truck door high-frequency details in view $I_S^{l r}$ than $I_1^{l r}$.  
    
    We unfold each warped depth map into patches $p_{i, j}$, $(i \in $ $ [1, S], j \in\left[1, \left(hr_h \times hr_w \right)/(ps^{2})\right])$,  $hr_h$, $hr_w$, and $ps$ denote the target HR height, target HR width, and patch size. 
    $j$ denotes the $j$-th patch index.
    We also unfold each warped view, and denote it as $I_{i \rightarrow T, j}^{h r}$.
    Then, we assign the depth value in each patch to its patch mean $p_{i, j}^{mean} = mean(p_{i, j})$.
    This alleviates the error of the depth value averaged in the edge area due to the up-sampling and warping operation.
    Then, we obtain the high-frequency index map $H$ from the index with the smallest depth value in $p_{i, j}^{mean}$ as follows:
    \begin{equation}
    {H}_{j}=\underset{i}{\arg \min } \ p_{i, j}^{mean}.
    \end{equation}
    $H_j$ refers to the high-frequency patch index of warped views, and we extract the high-frequency patch as the following formula:
    \begin{equation}
    {R}_{j}= I_{i \rightarrow T, j}^{h r}, \ i=H_j.
    \end{equation}
    Now we can obtain the multi-view reference image $R$ by folding the multi-view reference patches ${R}_{j}$.
    
    \noindent
    \textbf{Multiple References Synthesis.} 
    It is impossible to integrate all of the high-frequency details into a reference image. As a result, we synthesize $V$ multi-view reference images (MVRs) denoted as $\{R_i^{M}\}_{i=1}^V$ using those the top $V$ high-frequency index maps.
    However, the depth errors generated by MVS estimation and 3D warping causes inaccurate warped depth values, leading to incorrect MVR selection and alignment.
    In addition, the large angle difference between views magnifies alignment errors, resulting artifacts in MVR.
    Therefore, we synthesize $V$ near-view reference images (NVRs) $\{R_i^{N}\}_{i=1}^V$ from $L$ nearby views that have a small viewing angle gap and similar lighting conditions as the LR input view. NVRs suffer less from geometric inaccuracies, and the selection criteria for nearby views are based on their order in the dataset.
    NVR includes less high-frequency details than MVR, but it also contains fewer alignment errors, and artifacts.
    As a result, both MVRs and NVRs can be used to cross-reference to correct selection errors and accurately supply HF details to the SR model.

    \begin{figure*}[t]
    \centering
    \includegraphics[width=0.8\linewidth]{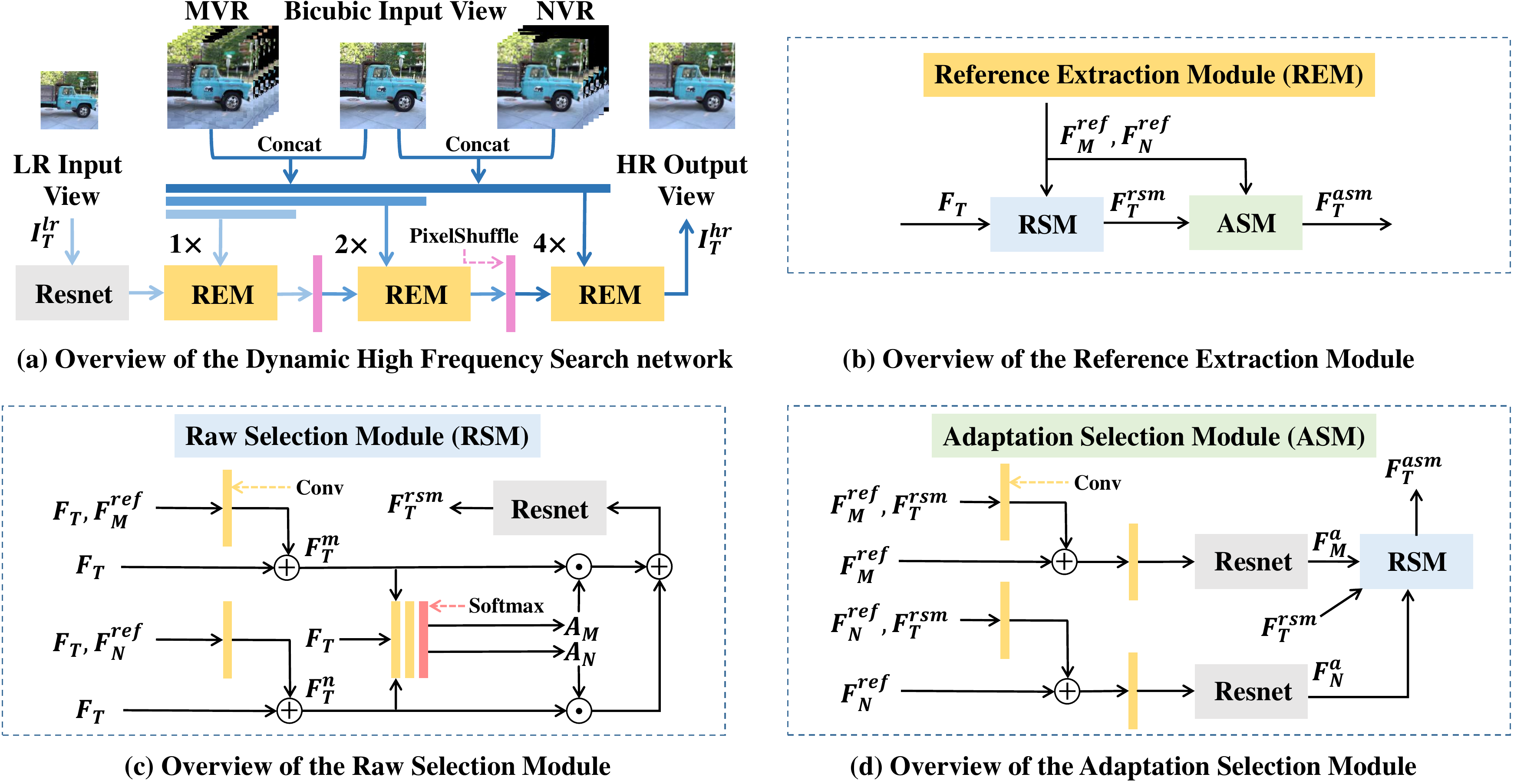}
    \caption{
    Overview of the proposed Dynamic High-Frequency Search (DHFS) network, which contains three Reference Extraction Module (REM). 
    Each REM includes a Raw Selection Module (RSM) and an Adaptation Selection Module (ASM) to exploit the high-frequency textural details in reference images dynamically.
    }
    \label{005_model_DHFS}
    \end{figure*}

\subsection{Dynamic High-Frequency Search}
\label{sec:DHFS}
    MVR has both more HF textural details and more artifacts than NVR, and that NVR is the inverse. Therefore, we propose the DHFS network to maximize the use of references (MVR and NVR).
    As shown in Figure~\ref{005_model_DHFS}(a), we first concatenate the MVRs and NVRs to the bicubic input view at the channel level, and use a learnable feature extractor to extract their multi-scale ($\times1$, $\times2$, and $\times4$) textural feature ($F_M^{ref}$ and $F_N^{ref}$).
    On the other hand, we obtain the textural features $F_T$ from LR input view using several residual layers. 
    Then, we use three Reference Extraction Module (REM) that do not share parameters. Each REM includes a Raw Selection Module (RSM) and an Adaptation Selection Module (ASM) to exploit the high-frequency textural details in reference images dynamically, as shown in Figure~\ref{005_model_DHFS}(b).
    We adopt PixelShuffle \cite{Shi_2016_CVPR} layer for up-sampling with factor 2.
    
    \noindent
    \textbf{Raw Selection Module.} RSM first extracts high-frequency textural details as residues from MVR features ($F_T,F_M^{ref}$) and NVR features ($F_T,F_M^{ref}$), and adds them to the feature of the LR input view $F_T$, getting $F_T^{m}$ and $F_T^{n}$ respectively.
    Then, it aggregates them into the final output $F_T^{rsm}$ via selection map $A_m$ and $A_n$.
    Figure~\ref{005_model_DHFS}(c) shows this process, and the formula is as follows:
    \begin{equation}
    \begin{aligned}
    &F_{T}^{m}=F_{T}+{Conv}_{1}\left(F_{T}, F_{M}^{r e f}\right)\\
    &F_{T}^{n}=F_{T}+{Conv}_{1}\left(F_{T}, F_{N}^{r e f}\right)\\
    &A_{M}, A_{N}={Softmax}\left({Conv}_{2}\left(F_{T}^{m}, F_{T}^{n}, F_{T}\right)\right) \\
    &F_{T}^{r s m}={Resnet}\left(A_{M} F_{T}^{m}+A_{N} F_{T}^{n}\right).
    \end{aligned}
    \label{equ:rsm}
    \end{equation}

    We find that RSM extracts more reference features from NVRs than from MVRs. 
    Reference features are generated by the feature extractor rather than residual layers such as $F_T$, resulting in a feature difference between reference feature and $F_T$.
    Due to the feature difference, it is difficult for the network to extract high-frequency textural details from MVR features which contain numerous artifacts and alignment errors.
    However, MVRs contain rich high-frequency textural details that are extremely useful for reconstructing the high-quality SR output view.
    As a result, we propose using ASM to reduce feature difference and feature artifacts, increasing DHFS capabilities of extracting MVR high-frequency details.


    \noindent
    \textbf{Adaptation Selection Module.}
    As shown in Figure~\ref{005_model_DHFS}(d), ASM first adds residues, calculated from ($F_M^{ref},F_T^{rsm}$) and ($F_N^{ref},F_T^{rsm}$), to $F_M^{ref}$ and $F_N^{ref}$.
    Then, utilizing numerous residual layers, ASM adapts reference features to $F_T^{rsm}$, reducing the feature difference and artifact features in reference features.
    Finally, it takes the adapted reference feature ($F_M^{a}$ and $F_N^{a}$) and $F_T^{rsm}$ as the input of another RSM to extract high-frequency feature details.
    We define the process in ASM as follows:
    \begin{equation}
    \begin{aligned}
        &F_{M}^{a}={Resnet}\left({Conv}_{2}\left(F_{M}^{r e f}+{Conv}_{1}\left(f_{M}^{r e f}, F_{T}^{r s m}\right)\right)\right)\\
        &F_{N}^{a}={Resnet}\left({Conv}_{2}\left(F_{N}^{r e f}+{Conv}_{1}\left(F_{N}^{r e f}, F_{T}^{r s m}\right)\right)\right)\\
        &F_{T}^{a s m}={RSM}\left({F}_{M}^{a}, F_{N}^{a}, F_{T}^{r s m}\right).
    \end{aligned}
    \label{equ:asm}
    \end{equation}
    As a result, dynamic extraction via RSM and ASM makes effective use of the HF textural details contained in the Refs.
    
     \begin{table*}[htb]
    \centering
    \resizebox{\textwidth}{!}
    { 
        \begin{tabular}{c|l|cccc|c}
        \hline
        & Model &Train & Playground & M60 & Truck & Total\\
        \hline
        \multirow{5}*{SISR}  &  EDSR \cite{Lim_2017_CVPR_Workshops} &  28.05 / 0.8294 / 0.226  &  28.33 / 0.8070 / 0.318  &  31.38 / 0.9072 / 0.165  &  28.19 / 0.8535 / 0.209  & 28.99 / 0.8493 / 0.230   \\ 
         &  RCAN \cite{zhang2018rcan} &  28.56 / 0.8351 / 0.228  &  29.68 / 0.8117 / 0.320  &  32.41 / 0.9126 / 0.163  &  28.93 / 0.8583 / 0.213  & 29.89 / 0.8544 / 0.231   \\ 
         &  HAN \cite{10.1007/978-3-030-58610-2_12} &  28.43 / 0.8321 / 0.222  &  29.59 / 0.8098 / 0.315  &  32.25 / 0.9103 / 0.167  &  28.77 / 0.8551 / 0.207  & 29.76 / 0.8518 / 0.228   \\ 
                  &  ESRGAN \cite{Wang_2018_ECCV_Workshops} &  25.53 / 0.7512 / 0.109  &  25.48 / 0.6810 / 0.151  &  29.52 / 0.8666 / 0.091  &  26.00 / 0.7893 / 0.097  & 26.63 / 0.7720 / 0.112   \\ 
                  &  SPSR \cite{Ma_2020_CVPR} &  25.68 / 0.7495 / 0.106  &  26.61 / 0.6900 / 0.133  &  29.85 / 0.8631 / 0.087  &  26.49 / 0.7920 / 0.090  & 27.16 / 0.7736 / 0.104   \\ 
        \hline
        \hline
        \multirow{4}*{VSR}   &  RBPN \cite{Haris_2019_CVPR} &  28.35 / 0.8314 / 0.236  &  29.66 / 0.8131 / 0.321  &  32.23 / 0.9117 / 0.171  &  28.77 / 0.8563 / 0.217  & 29.75 / 0.8531 / 0.236   \\ 
         &  RSDN \cite{10.1007/978-3-030-58610-2_38} &  27.68 / 0.8148 / 0.240  &  29.05 / 0.7954 / 0.314  &  31.05 / 0.8886 / 0.189  &  28.23 / 0.8402 / 0.218  & 29.00 / 0.8347 / 0.240   \\ 
         &  MuCAN \cite{10.1007/978-3-030-58607-2_20} &  28.46 / 0.8363 / 0.215  &  29.66 / 0.8148 / 0.296  &  32.56 / 0.9175 / 0.150  &  28.90 / 0.8607 / 0.200  & 29.90 / 0.8573 / 0.215   \\ 
         &  IconVSR \cite{Chan_2021_CVPR} &  28.31 / 0.8319 / 0.227  &  29.69 / 0.8172 / 0.310  &  32.36 / 0.9157 / 0.151  &  28.62 / 0.8538 / 0.218  & 29.75 / 0.8547 / 0.227   \\ 
        \hline
        \hline
       
                 \multirow{6}*{Ref-SR} & TTSR \cite{Yang_2020_CVPR} & 26.28 / 0.7661 / 0.118 & 27.08 / 0.7129 / 0.155 & 29.97 / 0.8626 / 0.108 & 26.80 / 0.8016 / 0.115 & 27.53 / 0.7858 / 0.124 \\ 
        & TTSR-rec \cite{Yang_2020_CVPR} & 28.16 / 0.8277 / 0.227 & 29.32 / 0.8039 / 0.319 & 31.84 / 0.9060 / 0.162 & 28.51 / 0.8509 / 0.214 & 29.46 / 0.8471 / 0.231 \\ 
        &     MASA \cite{Lu_2021_CVPR} &     25.48 / 0.7301 / 0.142 &     26.37 / 0.6822 / 0.152 &     28.89 / 0.8333 / 0.120 &     25.88 / 0.7722 / 0.121 &     26.66 / 0.7545 / 0.134 \\ 
        & MASA-rec \cite{Lu_2021_CVPR} & 28.43 / 0.8375 / 0.208 & 29.41 / 0.8073 / 0.309 & 32.10 / 0.9119 / 0.152 & 28.74 / 0.8586 / 0.205 & 29.67 / 0.8538 / 0.218 \\ 
                  &  $C^2$-Matching \cite{Jiang_2021_CVPR} &  28.13 / 0.8247 / 0.150  &  28.95 / 0.7844 / 0.210  &  31.71 / 0.9004 / 0.119  &  28.37 / 0.8441 / 0.151  & 29.29 / 0.8384 / 0.158   \\ 
         &  $C^2$-Matching-rec \cite{Jiang_2021_CVPR} &  28.71 / 0.8473 / 0.201  &  29.62 / 0.8160 / 0.305  &  32.32 / 0.9155 / 0.147  &  28.92 / 0.8648 / 0.194  & 29.89 / 0.8609 / 0.212   \\

        \hline
        \hline
        
                  &  MVSRnet  &  27.57 / 0.8222 / \textbf{0.085} & 27.99 / 0.7694 / \textbf{0.104} & 31.73 / 0.9065 / \textbf{0.068} & 28.03 / 0.8505 / \textbf{0.082} & 28.83 / 0.8372 / \textbf{0.085} \\
         &  MVSRnet-rec  &  \textbf{29.04} / \textbf{0.8588} / 0.183 & \textbf{29.93} / \textbf{0.8371} / 0.275 & \textbf{32.88} / \textbf{0.9278} / 0.127 & \textbf{29.19} / \textbf{0.8757} / 0.178 & \textbf{30.26} / \textbf{0.8748} / 0.191   \\ 
        \hline
        \end{tabular}
    }
    \caption{
    \textbf{Quantitative comparison on Tanks and Temples dataset.} $\uparrow$PSNR / $\uparrow$SSIM / $\downarrow$LPIPS are used for evaluation. SISR, VSR, and Ref-SR methods are grouped separately. ESRGAN, SPSR, TTSR, MASA, $C^2$-Matching, and MVSRnet are GAN-based methods and the best results are in bold.
    }
        \label{table:001_compare_sota_tanks}
    \end{table*}   
    
    \begin{table}[htb]
        \centering
        \resizebox{\linewidth}{!}
    { 
        \begin{tabular}{c|l|c|c}
        \hline
        & Model & BlendedMVS & GTAV \\
        \hline
        \multirow{5}*{SISR}  &  EDSR \cite{Lim_2017_CVPR_Workshops} & 27.80 / 0.7198 / 0.393 & 30.93 / 0.8478  / 0.211  \\ 
         &  RCAN \cite{zhang2018rcan} & 27.90 / 0.7227 / 0.393 & 31.23 / 0.8530  / 0.212 \\ 
         &  HAN \cite{10.1007/978-3-030-58610-2_12} & 27.86 / 0.7211 / 0.396 &  31.13 / 0.8515  / 0.206 \\ 
                  &  ESRGAN \cite{Wang_2018_ECCV_Workshops} & 25.50 / 0.6315 / 0.173 &  28.26 / 0.7746  / 0.105 \\ 
                  &  SPSR \cite{Ma_2020_CVPR} & 25.72 / 0.6406 / \textbf{0.168} &  28.37 / 0.7739  / 0.102 \\ 
        \hline
        \hline
        \multirow{4}*{VSR} &  RBPN \cite{Haris_2019_CVPR} & 27.79 / 0.7179 / 0.408 &  31.27 / 0.8559  / 0.218 \\ 
         &  RSDN \cite{10.1007/978-3-030-58610-2_38} & 26.38 / 0.6412 / 0.480 &  30.26 / 0.8365  / 0.223 \\ 
         &  MuCAN \cite{10.1007/978-3-030-58607-2_20} & 27.72 / 0.7162 / 0.400 &  31.24 / 0.8585  / 0.196 \\ 
         &  IconVSR \cite{Chan_2021_CVPR} & 27.53 / 0.7089 / 0.408 &  31.41 / 0.8633  / 0.205 \\ 
        \hline
        \hline
       
                 \multirow{7}*{Ref-SR} & TTSR \cite{Yang_2020_CVPR} & 26.09 / 0.6567 / 0.201 & 28.80 / 0.7914 / 0.121 \\ 
        & TTSR-rec \cite{Yang_2020_CVPR} & 27.68 / 0.7193 / 0.396 & 30.79 / 0.8485 / 0.216 \\ 
        &     MASA \cite{Lu_2021_CVPR} &      25.84 / 0.6468 / 0.220 &     28.05 / 0.7699 / 0.131 \\ 
        & MASA-rec \cite{Lu_2021_CVPR} & 27.82 / 0.7335 / 0.375 & 31.12 / 0.8613 / 0.196 \\ 
                 & $C^2$-Matching \cite{Jiang_2021_CVPR} & 27.24 / 0.7039 / 0.281 &  30.57 / 0.8396   / 0.153  \\ 
        & $C^2$-Matching-rec \cite{Jiang_2021_CVPR} & 27.99 / 0.7377 / 0.370 &  31.21 / 0.8628  / 0.194  \\ 
        \hline
        \hline
                  &  MVSRnet  & 26.90 / 0.7080 / 0.182 & 30.27 / 0.8462 / \textbf{0.085} \\ 
         &  MVSRnet-rec  & \textbf{28.03} /\textbf{ 0.7457} / 0.360  &  \textbf{31.60} / \textbf{0.8775} / 0.179 \\ 
        \hline
        \end{tabular}
    }
    \caption{
        \textbf{Quantitative comparison on BlendedMVS and GTAV datasets.} $\uparrow$PSNR / $\uparrow$SSIM / $\downarrow$LPIPS are used for evaluation. 
        SISR, VSR, and Ref-SR methods are grouped separately.
    }
        \label{table:002_compare_blendedmvs_gtav}
    \end{table}

\subsection{Loss Functions}
\label{sec:loss}
    Similar to most SR models, we adopt the reconstruction loss $\mathcal{L}_{{rec}}$, adversarial loss $\mathcal{L}_{{adv}}$, and perceptual loss $\mathcal{L}_{{per}}$ in our training stage. The  $\mathcal{L}_{{adv}}$ and $\mathcal{L}_{{per}}$ are used to improve the visual quality, and we define the overall loss as follows:
    \begin{equation}
        \mathcal{L}_{{overall }}=\mathcal{L}_{{rec }}+\lambda_{a d v} \mathcal{L}_{a d v}+\lambda_{{per }} \mathcal{L}_{{per }}
    \label{equ:overall_loss}
    \end{equation}
    Specifically, we employ $\ell_{1}$-norm for reconstruction loss and adopt the WGAN-GP \cite{DBLP:conf/nips/GulrajaniAADC17} structure to obtain adversarial loss.
    For perceptual loss \cite{10.1007/978-3-319-46475-6_43}, it is calculated on the VGG19 \cite{Simonyan15} features of conv3-2, conv4-2 ,and conv5-2.

\section{Experiments}

\subsection{Datasets and Training Details}
\noindent
\textbf{Datasets.} As the multi-view image super-resolution problem focuses on multi-view settings, we compare our MVSRnet to the state of the art on three challenging multi-view datasets: Tanks and Temples \cite{DBLP:journals/tog/KnapitschPZK17}, BlendedMVS \cite{Yao_2020_CVPR}, and GTAV \cite{Huang_2018_CVPR}.
The real-world images captured in Tanks and Temples dataset and BlendedMVS are irregular distribution.
GTAV is a photo-realistic synthetic dataset containing urban scenes captured in the video game Grand Theft Auto V.
Specifically, we use 17 scenes from Tanks and Temples for training and 4 (Train, Playground, M60, and Truck) for testing, the same as \cite{Riegler2020FVS}.
In addition, the camera pose and the depth map we use are provided by \cite{Riegler2020FVS} for Tanks and Temple. 
Additionally, we evaluate ten scenes in BlendedMVS and all scenes in GTAV, proving the generalization capacity of our MVSRnet.
The number of views in each scene of the Tanks and Temples and BlendedMVS datasets is concentrated in 300-400 and 60-150, and the GTAV dataset is 100.
We set the SR factor as $\times4$, and we generate LR multi-view images and depth maps by bicubic and modify the camera intrinsic matrices according to the SR factor.

    \begin{figure*}[t]
    \centering
    \includegraphics[width=0.95\linewidth]{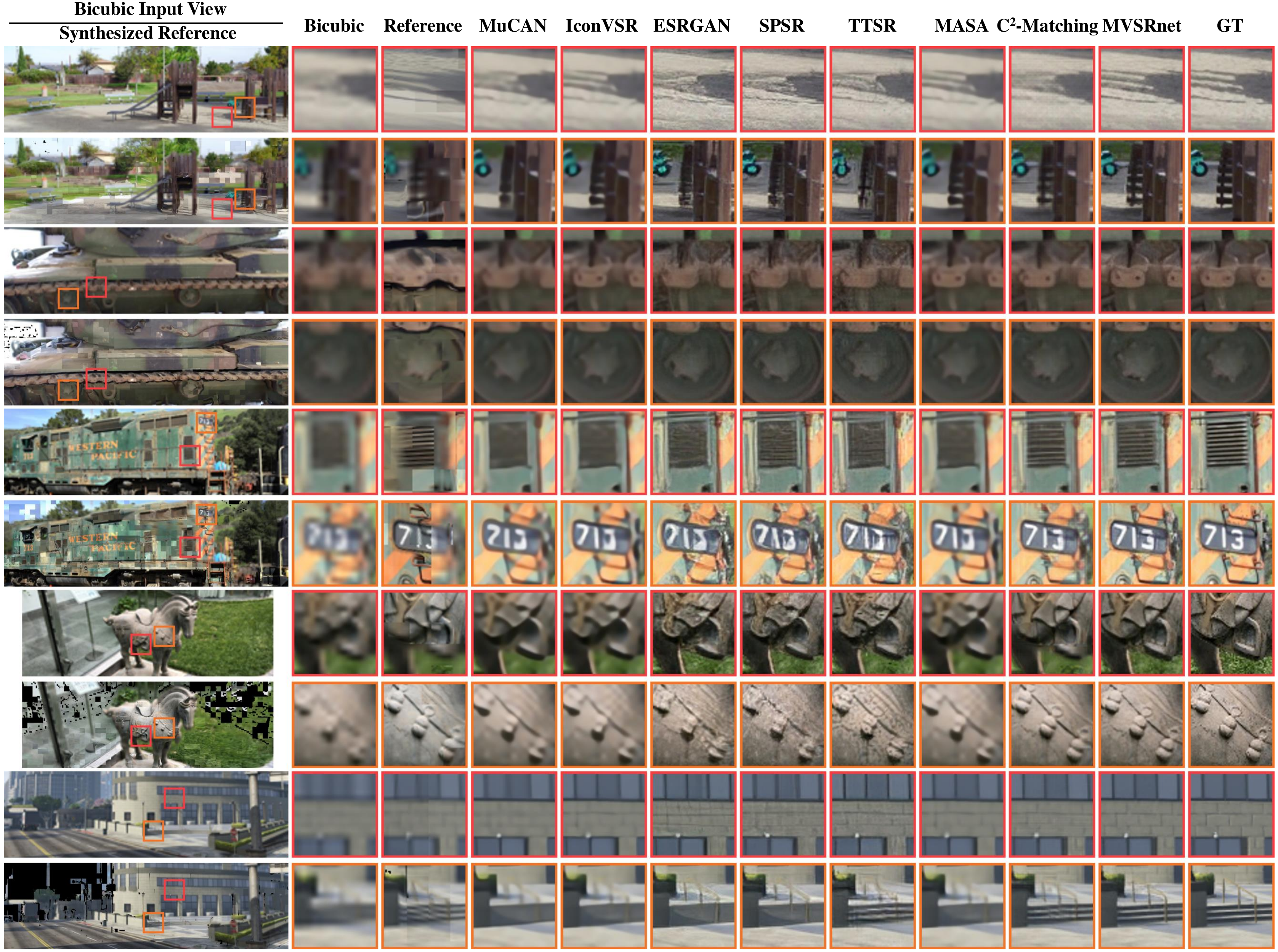}
    \caption{
    \textbf{ Qualitative comparisons.}
    The top three examples are Playground, M60, and Train scene in Tanks and Temple dataset. The fourth and fifth example come from BlenedMVS and GTAV datasets, respectively. 
    Zoom in for better visualization.
    }
    \label{006_experiment_001}
    \end{figure*}

\noindent
\textbf{Training Details.} 
We first utilize the Geometry-Aware Reference Synthesis module to synthesize MVRs and NVRs for each LR input view. 
Specifically, we synthesize $V$ MVRs and $V$ NVRs from all LR multi-view images and $L$ adjacent views, respectively. 
$V$ is equal to 6, and we empirically set the $L$ to 6 .
Then, we use synthesized reference images and the LR input view to train the Dynamic High-Frequency Search network from scratch on an Nvidia GeForce RTX 3090 GPU.
We implement our method in Pytorch, and the total number of training iterations is 200,000, with a batch size of 2 and a LR patch size of $80\times160$ pixels.
In addition, we decay the learning rate from 1e-4 to 1e-7 in a cosine annealing way, and we choose ADAM with parameters $\beta_1 = 0.9$, and $\beta_2 = 0.999$ as optimizer. We set $\lambda_{a d v}$ and $\lambda_{{per }}$ in Eq.(\ref{equ:overall_loss}) to 5e-3 and 1e-2.

\subsection{Comparison with State-of-the-Art Methods}

We compare our MVSRnet with recent SISR, VSR, and Ref-SR methods. 
We use five SISR methods, including EDSR \cite{Lim_2017_CVPR_Workshops}, RCAN \cite{zhang2018rcan}, HAN \cite{10.1007/978-3-030-58610-2_12}, ESRGAN \cite{Wang_2018_ECCV_Workshops}, and SPSR \cite{Ma_2020_CVPR}, four VSR methods, including RBPN \cite{Haris_2019_CVPR}, RSDN \cite{10.1007/978-3-030-58610-2_38}, MuCAN \cite{10.1007/978-3-030-58607-2_20}, and IconVSR \cite{Chan_2021_CVPR}, and three Ref-SR methods, including TTSR \cite{Yang_2020_CVPR}, MASA \cite{Lu_2021_CVPR}, and $C^2$-Matching \cite{Jiang_2021_CVPR}. 
We use the best results from their pre-trained and fine-tuned models using our Tanks and Temples dataset for display.
Because RSDN and MuCAN do not supply training code, we rely on the pre-trained models they provide.
We employ an MVR synthesized from the top-1 HF index maps as the input to Ref-SR methods because these methods are incapable of processing hundreds of reference images.
Our compared Ref-SR methods may concatenate the references in a row or column to conduct multi-reference SR, but the GPU runs out of memory when employing two synthesized references.
ESRGAN, SPSR, TTSR, MASA, and $C^2$-Matching are all GAN-based models, whereas the remaining models rely solely on reconstruction loss.
To ensure a fair comparison, we train our model with overall loss (Eq.(\ref{equ:overall_loss})) and solely reconstruction loss and refer to them as MVSRnet and MVSRnet-rec, respectively.

\noindent
\textbf{Quantitative Comparison.}
Table~\ref{table:001_compare_sota_tanks} presents the quantitative results of the Tanks and Temples dataset. 
In Table~\ref{table:002_compare_blendedmvs_gtav}, we also verify the generalization capacity of our MVSRnet using BlendedMVS and GTAV, and our approach continues to perform the best.
The PSNR of GAN-based $C^2$-Matching in Table~\ref{table:001_compare_sota_tanks} and Table~\ref{table:002_compare_blendedmvs_gtav} is higher than our MVSRnet, while MVSRnet reduces the LPIPS \cite{Zhang_2018_CVPR} error by around 35\%-46\% when compared to $C^2$-Matching.
LPIPS is validated to human perception and is commonly used as a metric for evaluating GAN-based models.
While our LPIPS is comparable to that of SPSR in BlendedMVS, our PSNR and SSIM are 1.18dB and 0.0674 higher, respectively.
In summary, our proposed method outperforms the state of the art by a significant margin.

\noindent
\textbf{Qualitative Evaluation.}
The visual results shown in Figure~\ref{006_experiment_001} reveal that our method is capable of producing finer details, and they lead to two conclusions.
(a) Our reference image generated from all LR multi-view images is more beneficial for SR than the image self-reference (SISR) and adjacent views (VSR). For example, the reference image of the fourth example contains more realistic horse details than the bicubic input view. Thus, using our reference image, MVSRnet generates more fine textural details of the horse when compared to SISR methods (ESRGAN and SPSR) and VSR methods (MuCAN and IconVSR). 
(b) Our DHFS network, including RSM and ASM, shows its effectiveness and robustness in identifying and extracting useful features from the reference images. Specifically, because our synthesized reference image is used as an input, Ref-SR methods such as TTSR, MASA, and $C^2$-Matching can produce textural results. They are, however, not robust to artifacts in the synthesized reference.
For example, in Figure~\ref{006_experiment_001}, the reference image of the third example contains blur artifacts, resulting in more blur areas in the $C^2$-Matching image than in our MVSRnet image.

    \begin{figure*}[t]
    \centering
    \includegraphics[width=\linewidth]{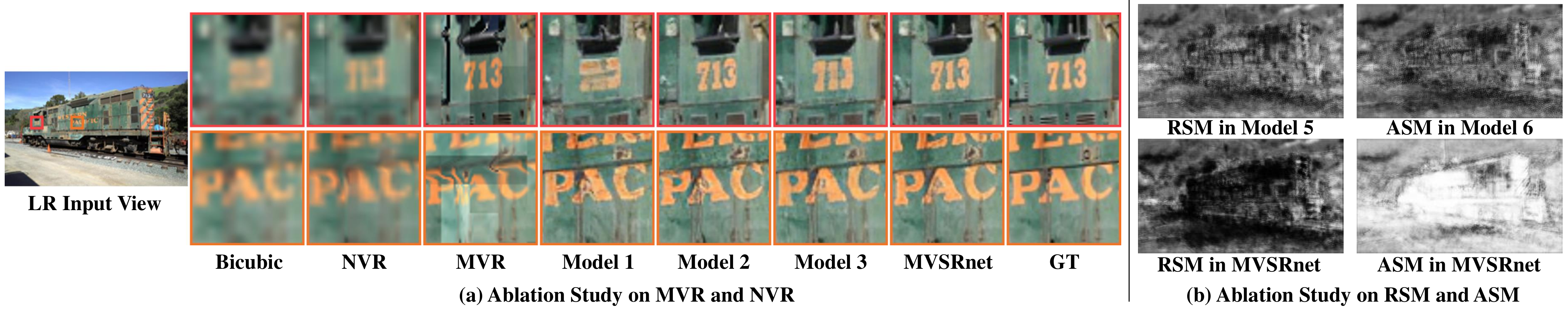}
    \caption{
    \textbf{Ablation study.} (a) Effectiveness of MVR and NVR. Model 1 uses neither MVR nor NVR, and Model 2 and 3 only use MVR or NVR. (b) Effectiveness of RSM and ASM. The upper left and upper right are the selection map $A_M$ of features $F_T^{M}$ from RSM in Model 5 and from ASM in Model 6, respectively. The lower left and lower right are the selection map $A_M$ from RSM and from ASM in MVSRnet.
    }
    \label{008_experiment_004}
    \end{figure*}

\subsection{Ablation Study}

\noindent
\textbf{Muti-View Reference Image (MVR) and Near-View Reference Image (NVR)}.
We analyze the MVR and NVR that have complementary effects on SR tasks. The ablation Model 1 does not input both MVR and NVR, and Model 2 and Model 3 do not input NVR and MVR, respectively.
As illustrated in Figure~\ref{008_experiment_004}(a), MVR contains high-frequency details (sharp number `713') as well as noticeable artifacts (the incorrect structure of letter `PAC').
In contrast, NVR has fewer artifacts and details of `PAC' and `713' than MVR, but there are still more high-frequency details than the bicubic input view.
We discover that MVSRnet recovers the exact number `713' than Model 3 and produces fewer `PAC' artifacts than Model 2.
Additionally, we conduct quantitative ablation study, and Table~\ref{table:003_ablation} reveals that MVSRnet has superior PSNR performance than Model 2 and Model 3 by 0.38dB and 0.45 dB, respectively.
These experiments fully demonstrate the complementary effects of MVR and NVR.

In addition, we search for the optimal number $V$ of MVR and NVR.
The LR details and artifacts areas grow in relation to the number $V$ of reference images, making it challenging for the network to identify and extract relevant high frequencies details.
We perform ablation studies with $V$ to 1, 2, 4, 6, and 8, and we achieve the best performance when we set $V$ to 6, as shown in Table~\ref{table:optimal_number}. 
We also conduct an ablation study on the hyper-parameter of patch size $ps$ as 4/8/16/32/64 in GARS module, and achieve the best performance when we set $ps$ to 16, as shown in Table~\ref{table:optimal_number}. The reason is that when the patch size is too small, there are numerous selection mistakes and block-boundary artifacts in the synthesized reference image. When the block size is too large, the selected patches are blurry and lack a significant amount of high-frequency texture details.



\noindent
\textbf{Raw Selection Module (RSM) and Adaptation Selection Module (ASM)}.
The ablation Model 4 does not include both RSM and ASM, and Model 5 and Model 6 do not include ASM and RSM, respectively.
Figure~\ref{008_experiment_004}(b) shows that selected areas (denoted as white) in the selection map of `ASM in MVSRnet' are more than `RSM in Model 5' and `ASM in Model 6', indicating that more MVR HR feature details are selected for SR.
This is because a small difference between features can help the network identify and extract useful features.
As mentioned in Section~\ref{sec:DHFS}, ASM implicitly decreases the difference between reference features and input view features and reduces artifact features in reference features, making it easier for the network to extract useful features in MVR.
RSM, on the other hand, uses a residual process to adapt input view features to reference features, as indicated in Eq.(\ref{equ:rsm}).
The different ways ASM and RSM incorporate features makes feature extraction more sufficient.
Thus, combining RSM and ASM significantly improves the power of DHFS to extract and identify HF details when compared to using only RSM or ASM.
We also verify them using the quantitative evaluation displayed in Table~\ref{table:003_ablation}.
MVSRnet improves the PSNR value by 0.12dB and 0.11dB, respectively, when compared to Model 5 and 6, demonstrating the usefulness of using both RSM and ASM.

     \begin{table}[t]
        \centering
        \resizebox{\linewidth}{!}
    { 
        \begin{tabular}{c|c|c|c|c|c|c}
        \hline
        Model & MVR & NVR & RSM & ASM & Parameters & Tanks and Temple \\
        \hline
        Model 1 &  & & \Checkmark  & \Checkmark & 19.51M & 27.33 / 0.7734 / 0.108 \\
        Model 2 & \Checkmark & & \Checkmark  & \Checkmark & 19.56M & 28.45 / 0.8215 / 0.089 \\
        Model 3 &  & \Checkmark & \Checkmark  & \Checkmark & 19.56M & 28.38 / 0.8182 / 0.089 \\
        \hline
        Model 4 & \Checkmark  & \Checkmark &  &  & 19.73M & 28.68 / 0.8323 / 0.085 \\
        Model 5 & \Checkmark  & \Checkmark & \Checkmark &  & 19.62M & 28.71 / 0.8324 / 0.084 \\
        Model 6 & \Checkmark  & \Checkmark &  & \Checkmark & 19.46M & 28.72 / 0.8340 / \textbf{0.083} \\
        \hline
        MVSRnet & \Checkmark  & \Checkmark & \Checkmark  & \Checkmark & 19.42M & \textbf{28.83} / \textbf{0.8372} / \textbf{0.085} \\
    
        \hline
        \end{tabular}
    }
        \caption{
        Quantitative ablation study on the MVR and NVR and quantitative ablation study on the RSM and ASM.
        }
        \label{table:003_ablation}
    \end{table}

 \begin{table}[t]
    \centering
    \resizebox{\linewidth}{!}
{ 

    \begin{tabular}{c|c||c|c}
    \hline
    $V$ & Tanks \& Temple & Patch Size  & Tanks \& Temple \\
    \hline
    1 & 28.49 / 0.8252 / 0.087 &
    4 & 28.72 / 0.8335 / \textbf{0.082} \\
    2 & 28.73 / 0.8353 / \textbf{0.085} &
    8 & 28.76 / 0.8336 / 0.084 \\
    4 & 28.71 / 0.8334 / \textbf{0.085} &
    16 & \textbf{28.83} / \textbf{0.8372} / 0.085 \\
     6 & \textbf{28.83} / \textbf{0.8372} / \textbf{0.085} &
    32 & 28.73 / 0.8323 / 0.086 \\
    8 & 28.56 / 0.8289 / 0.087 & 64 & 28.56 / 0.8273 / 0.087 \\
    \hline
    \end{tabular}
}
    \caption{Columns 1 and 2: ablation study of the optimal number $V$ of MVR and $V$ NVR. Columns 3 and 4: ablation study of the patch size $ps$ in GARS module.
    }
    \label{table:optimal_number}
\end{table}



\section{Conclusion}
We propose the MVSRnet for multi-view image super-resolution that aims to super-resolve each view by utilizing all views as references.
Our Geometry-Aware Reference Synthesis module, in particular, can synthesize reference images from all LR multi-view images by explicitly using geometry information.
The reference image contains high-frequency details and is aligned with the input view.
Then we implement a Dynamic High-Frequency Search network which includes the Raw Selection Module and Adaptation Selection Module, to dynamically select useful features in the synthesized reference features for super-resolution.
We have achieved best experimental results compared with state-of-the-art methods and this work enriches the Reference-Based SR research.

\bibliographystyle{ACM-Reference-Format}
\bibliography{sample-base}

\appendix

\section{Loss Functions}
\label{loss_functions}

\noindent
\textbf{Reconstruction Loss.} The $\ell_{1}$-norm of the reconstruction loss is used as follows:
\begin{equation}
\mathcal{L}_{\text {rec }}=\left\|I_T^{h r}-I_T^{g t}\right\|_{1},
\label{l1}
\end{equation}
where $I_T^{h r}$ and $I_T^{g t}$ represent the output of our MVSRnet and the high-resolution (HR) view of the ground truth, respectively.

\noindent
\textbf{Adversarial Loss.}
We use the WGAN-GP \cite{DBLP:conf/nips/GulrajaniAADC17} structure to obtain adversarial loss \cite{DBLP:conf/nips/GoodfellowPMXWOCB14}.  
WGAN-GP replaces the weight clipping with a gradient norm penalty to achieve stable training and generate high-quality samples. We define adversarial loss as follows:
\begin{equation}
\begin{aligned}
\mathcal{L}_{D}=& \mathbb{E}_{\boldsymbol{I_T^{hr}}}[D(I_T^{hr})]-\mathbb{E}_{\boldsymbol{I_T^{gt}}}[D(I_T^{gt})]+\\
& \lambda {\mathbb{E}_{\boldsymbol{\hat{I}}}}\left[\left(\left\|\nabla_{\hat{I}} D(\hat{I})\right\|_{2}-1\right)^{2}\right],
\end{aligned}
\end{equation}
\begin{equation}
\mathcal{L}_{G}=-{\mathbb{E}_{\boldsymbol{I_T^{hr}}}}[D(I_T^{hr})].
\end{equation}
where $\hat{I}$ is the random sample of our network output $I_T^{h r}$ and ground truth $I_T^{g t}$ through combination, ($\hat{I} \leftarrow \epsilon I_T^{gt}+(1-\epsilon) I_T^{h r}, \epsilon \sim U[0,1]$).

\noindent
\textbf{Perceptual Loss.} We utilize perceptual loss \cite{10.1007/978-3-319-46475-6_43} to generate the photo-realist image, which we define as follows:

\begin{equation}
L_{p e r}= \left\|\phi_{i}\left(I_T^{h r}\right)-\phi_{i}\left(I_T^{g t}\right)\right\|_{1},
\end{equation}
where $\phi_{i}$ denotes the $i$-th layer feature map of VGG19 \cite{Simonyan15}, and we use the features of conv3-2, conv4-2, and conv5-2 to calculate the perceptual loss.

\begin{table}[h]
    \caption{
        \textbf{Comparison to multi-view super-resolution method.}
    }
        \centering
        \resizebox{\linewidth}{!}
        { 
            \begin{tabular}{l|c|c|c}
        \hline
         Model & Tanks and Temple & BlendedMVS \\
        \hline
        SASRnet \cite{DBLP:conf/mm/SunC0Z21} &  28.99 / 0.8493 / 0.230 & 27.80 / 0.7198 / 0.393 \\ 
        MVSRnet-rec & \textbf{30.26} / \textbf{0.8748} / \textbf{0.191}  & \textbf{28.03} / \textbf{0.7457} / \textbf{0.360} \\ 
         
        \hline
        \end{tabular}
        }

        \label{tab:compare_mvsr}
    \end{table}

\begin{table*}[t]
    \caption{\textbf{The network structure of our Dynamic High-Frequency Search network.} MVR, NVR, and Bic denote the multi-view reference image, near-view reference image, and the bicubic of the LR input view, respectively. We present the convolution layer as Conv(input channels, output channels, kernel size, stride, padding). $RB\times N$ is a residual block with $N$ residual layers, and we adopt the residual layer with channel attention described in \cite{zhang2018rcan}. In addition, we use PixelShuffle to upsample, and use ReLU in $RB$ and LeakyReLU other places. 
    }
    \centering
    \resizebox{\textwidth}{!}
    {
        \begin{tabular}{|c|c|c|c|c|c|c|c|c|c|c|c|}
        \hline
         & Id & Layer Name &  & Id & Layer Name &  & Id & Layer Name &  & Id & \begin{tabular}[c]{@{}c@{}}Layer\\ Name\end{tabular} \\ \hline
        \multirow{8}{*}{\begin{tabular}[c]{@{}c@{}}Feature\\ Extractor\end{tabular}} & 1-0 & Concat (MVRs, Bic) & \multirow{2}{*}{} & 2-0 & LR Input View & \multirow{2}{*}{} & \multirow{2}{*}{} & \multirow{2}{*}{} & \multirow{2}{*}{} & \multirow{2}{*}{} & \multirow{2}{*}{} \\ \cline{2-3} \cline{5-6}
         & 1-1 & \begin{tabular}[c]{@{}c@{}}Conv(19,64,3,1,1),\\ RB $\times$  3\end{tabular} &  & 2-1 & \begin{tabular}[c]{@{}c@{}}Conv(3,64,3,1,1),\\ RB $\times$ 10\end{tabular} &  &  &  &  &  &  \\ \cline{2-12} 
         & 1-2 & \begin{tabular}[c]{@{}c@{}}Conv(64,128,4,2,1),\\ RB $\times$ 3\end{tabular} & \multirow{6}{*}{\begin{tabular}[c]{@{}c@{}}RSM\\ (scale 1$\times$)\end{tabular}} & 2-2 & \begin{tabular}[c]{@{}c@{}}Concat(\$2-1, \$1-3),\\ Conv(320,64,3,1,1)\end{tabular} & \multirow{6}{*}{\begin{tabular}[c]{@{}c@{}}RSM\\ (scale 2$\times$)\end{tabular}} & 3-0 & \begin{tabular}[c]{@{}c@{}}Concat(\$2-18, \$1-2),\\ Conv(192,64,3,1,1)\end{tabular} & \multirow{6}{*}{\begin{tabular}[c]{@{}c@{}}RSM\\ (scale 4$\times$)\end{tabular}} & 4-0 & \begin{tabular}[c]{@{}c@{}}Concat(\$3-16, \$1-1),\\ Conv(128,64,3,1,1)\end{tabular} \\ \cline{2-3} \cline{5-6} \cline{8-9} \cline{11-12} 
         & 1-3 & \begin{tabular}[c]{@{}c@{}}Conv(128,256,4,2,1),\\ RB $\times$ 3\end{tabular} &  & 2-3 & \$2-1+\$2-2 &  & 3-1 & \$2-18+\$3-0 &  & 4-1 & \$3-16+\$4-0 \\ \cline{2-3} \cline{5-6} \cline{8-9} \cline{11-12} 
         & 1-4 & Concat (NVRs, Bic) &  & 2-4 & \begin{tabular}[c]{@{}c@{}}Concat(\$2-1, \$1-7),\\ Conv(320,64,3,1,1)\end{tabular} &  & 3-2 & \begin{tabular}[c]{@{}c@{}}Concat(\$2-18, \$1-6),\\ Conv(192,64,3,1,1)\end{tabular} &  & 4-2 & \begin{tabular}[c]{@{}c@{}}Concat(\$3-16, \$1-5),\\ Conv(128,64,3,1,1)\end{tabular} \\ \cline{2-3} \cline{5-6} \cline{8-9} \cline{11-12} 
         & 1-5 & \begin{tabular}[c]{@{}c@{}}Conv(19,64,3,1,1),\\ RB $\times$ 3\end{tabular} &  & 2-5 & \$2-1+\$2-4 &  & 3-3 & \$2-18+\$3-2 &  & 4-3 & \$3-16+\$4-2 \\ \cline{2-3} \cline{5-6} \cline{8-9} \cline{11-12} 
         & 1-6 & \begin{tabular}[c]{@{}c@{}}Conv(64,128,4,2,1),\\ RB $\times$ 3\end{tabular} &  & 2-6 & \begin{tabular}[c]{@{}c@{}}Concat(\$2-3,\$2-5,\$2-1),\\ Conv(192,64,3,1,1),\\ LeakyReLU,\\ Conv(64,2,3,1,1),\\ Softmax\end{tabular} &  & 3-4 & \begin{tabular}[c]{@{}c@{}}Concat(\$3-1,\$3-3,\$2-18),\\ Conv(192,64,3,1,1),\\ LeakyReLU,\\ Conv(64,2,3,1,1),\\ Softmax\end{tabular} &  & 4-4 & \begin{tabular}[c]{@{}c@{}}Concat(\$4-1,\$4-3,\$3-16),\\ Conv(192,64,3,1,1),\\ LeakyReLU,\\ Conv(64,2,3,1,1),\\ Softmax\end{tabular} \\ \cline{2-3} \cline{5-6} \cline{8-9} \cline{11-12} 
         & 1-7 & \begin{tabular}[c]{@{}c@{}}Conv(128,256,4,2,1),\\ RB $\times$ 3\end{tabular} &  & 2-7 & \begin{tabular}[c]{@{}c@{}}\$2-3$\times$\$2-6[0]+\\ \$2-5$\times$\$2-6[1],\\ RB $\times$ 8\end{tabular} &  & 3-5 & \begin{tabular}[c]{@{}c@{}}\$3-1$\times$\$3-4[0]+\\ \$3-3$\times$\$3-4[1],\\ RB $\times$ 6\end{tabular} &  & 4-5 & \begin{tabular}[c]{@{}c@{}}\$4-1$\times$\$4-4[0]+\\ \$4-3$\times$\$4-4[1],\\ RB $\times$ 4\end{tabular} \\ \hline
        \multirow{11}{*}{} & \multirow{11}{*}{} & \multirow{11}{*}{} & \multirow{11}{*}{\begin{tabular}[c]{@{}c@{}}ASM\\ (scale 1$\times$)\end{tabular}} & 2-8 & \begin{tabular}[c]{@{}c@{}}Concat(\$1-3, \$2-7),\\ Conv(320,256,3,1,1)\end{tabular} & \multirow{11}{*}{\begin{tabular}[c]{@{}c@{}}ASM\\ (scale 2$\times$)\end{tabular}} & 3-6 & \begin{tabular}[c]{@{}c@{}}Concat(\$1-2, \$3-5),\\ Conv(192,128,3,1,1)\end{tabular} & \multirow{11}{*}{\begin{tabular}[c]{@{}c@{}}ASM\\ (scale 4$\times$)\end{tabular}} & 4-6 & \begin{tabular}[c]{@{}c@{}}Concat(\$1-1, \$4-5),\\ Conv(128,64,3,1,1)\end{tabular} \\ \cline{5-6} \cline{8-9} \cline{11-12} 
         &  &  &  & 2-9 & \begin{tabular}[c]{@{}c@{}}\$1-3+\$2-8,\\ Conv(256,64,3,1,1),\\ RB $\times$ 8\end{tabular} &  & 3-7 & \begin{tabular}[c]{@{}c@{}}\$1-2+\$3-6,\\ Conv(128,64,3,1,1),\\ RB $\times$ 6\end{tabular} &  & 4-7 & \begin{tabular}[c]{@{}c@{}}\$1-1+\$4-6,\\ Conv(64,64,3,1,1),\\ RB $\times$ 4\end{tabular} \\ \cline{5-6} \cline{8-9} \cline{11-12} 
         &  &  &  & 2-10 & \begin{tabular}[c]{@{}c@{}}Concat(\$1-7, \$2-7),\\ Conv(320,256,3,1,1)\end{tabular} &  & 3-8 & \begin{tabular}[c]{@{}c@{}}Concat(\$1-6, \$3-5),\\ Conv(192,128,3,1,1)\end{tabular} &  & 4-8 & \begin{tabular}[c]{@{}c@{}}Concat(\$1-5, \$4-5),\\ Conv(64,64,3,1,1)\end{tabular} \\ \cline{5-6} \cline{8-9} \cline{11-12} 
         &  &  &  & 2-11 & \begin{tabular}[c]{@{}c@{}}\$1-7+\$2-10,\\ Conv(256,64,3,1,1)\\ RB $\times$ 8\end{tabular} &  & 3-9 & \begin{tabular}[c]{@{}c@{}}\$1-6+\$3-8,\\ Conv(128,64,3,1,1)\\ RB $\times$ 6\end{tabular} &  & 4-9 & \begin{tabular}[c]{@{}c@{}}\$1-5+\$4-8,\\ Conv(64,64,3,1,1)\\ RB $\times$ 4\end{tabular} \\ \cline{5-6} \cline{8-9} \cline{11-12} 
         &  &  &  & 2-12 & \begin{tabular}[c]{@{}c@{}}Concat(\$2-7, \$2-9),\\ Conv(128,64,3,1,1)\end{tabular} &  & 3-10 & \begin{tabular}[c]{@{}c@{}}Concat(\$3-5, \$3-7),\\ Conv(128,64,3,1,1)\end{tabular} &  & 4-10 & \begin{tabular}[c]{@{}c@{}}Concat(\$4-5, \$4-7),\\ Conv(128,64,3,1,1)\end{tabular} \\ \cline{5-6} \cline{8-9} \cline{11-12} 
         &  &  &  & 2-13 & \$2-7+\$2-12 &  & 3-11 & \$3-5+\$3-10 &  & 4-11 & \$4-5+\$4-10 \\ \cline{5-6} \cline{8-9} \cline{11-12} 
         &  &  &  & 2-14 & \begin{tabular}[c]{@{}c@{}}Concat(\$2-7, \$2-11),\\ Conv(128,64,3,1,1)\end{tabular} &  & 3-12 & \begin{tabular}[c]{@{}c@{}}Concat(\$3-5, \$3-9),\\ Conv(128,64,3,1,1)\end{tabular} &  & 4-12 & \begin{tabular}[c]{@{}c@{}}Concat(\$4-5, \$4-9),\\ Conv(128,64,3,1,1)\end{tabular} \\ \cline{5-6} \cline{8-9} \cline{11-12} 
         &  &  &  & 2-15 & \$2-7+\$2-14 &  & 3-13 & \$3-5+\$3-12 &  & 4-13 & \$4-5+\$4-12 \\ \cline{5-6} \cline{8-9} \cline{11-12} 
         &  &  &  & 2-16 & \begin{tabular}[c]{@{}c@{}}Concat(\$2-13,\$2-15,\$2-7),\\ Conv(192,64,3,1,1),\\ LeakyReLU,\\ Conv(64,2,3,1,1),\\ Softmax\end{tabular} &  & 3-14 & \begin{tabular}[c]{@{}c@{}}Concat(\$3-11,\$3-13,\$3-5),\\ Conv(192,64,3,1,1),\\ LeakyReLU,\\ Conv(64,2,3,1,1),\\ Softmax\end{tabular} &  & 4-14 & \begin{tabular}[c]{@{}c@{}}Concat(\$4-11,\$4-13,\$4-5),\\ Conv(192,64,3,1,1),\\ LeakyReLU,\\ Conv(64,2,3,1,1),\\ Softmax\end{tabular} \\ \cline{5-6} \cline{8-9} \cline{11-12} 
         &  &  &  & 2-17 & \begin{tabular}[c]{@{}c@{}}\$2-13$\times$\$2-16[0]+\\ \$2-15$\times$\$2-16[1],\\ RB $\times$ 8\end{tabular} &  & 3-15 & \begin{tabular}[c]{@{}c@{}}\$3-11$\times$\$3-14[0]+\\ \$3-13$\times$\$3-14[1],\\ RB $\times$ 6\end{tabular} &  & 4-15 & \begin{tabular}[c]{@{}c@{}}\$4-11$\times$\$4-14[0]+\\ \$4-13$\times$\$4-14[1],\\ RB $\times$ 4\end{tabular} \\ \cline{5-6} \cline{8-9} \cline{11-12} 
         &  &  &  & 2-18 & \begin{tabular}[c]{@{}c@{}}Conv(64,256,3,1,1),\\ PixelShuffle(2),\\ Conv(64,64)\end{tabular} &  & 3-16 & \begin{tabular}[c]{@{}c@{}}Conv(64,256,3,1,1),\\ PixelShuffle(2),\\ Conv(64,64)\end{tabular} &  & 4-16 & Conv(64,3,3,1,1), \\ \hline
        \end{tabular}

    }
    \label{tab:dhfs}
\end{table*}

\begin{table*}[t]
    \caption{
        \textbf{Comparison of the model size.}
        \emph{light} denotes the light version of our proposed network.
        \textcolor[rgb]{1,0,0}{Red} and \textcolor[rgb]{0,0,1}{blue} indicate first and second performance, respectively. We include the parameter of discriminator in the total parameters of the GAN-based model.
    }
        \centering
        \resizebox{\linewidth}{!}
        { 
            \begin{tabular}{c|l|c|c|c|c}
        \hline
        & Model & Parameters & Tanks and Temple & BlendedMVS & GTAV \\
        \hline
        \multirow{5}*{SISR}  &  EDSR \cite{Lim_2017_CVPR_Workshops} & 43.09M  & 28.99 / 0.8493 / 0.230 & 27.80 / 0.7198 / 0.393 & 30.93 / 0.8478  / 0.211  \\ 
         &  RCAN \cite{zhang2018rcan} & 15.59M & 29.89 / 0.8544 / 0.231 & 27.90 / 0.7227 / 0.393 & 31.23 / 0.8530  / 0.212 \\ 
         &  HAN \cite{10.1007/978-3-030-58610-2_12} & 16.07M & 29.76 / 0.8518 / 0.228 & 27.86 / 0.7211 / 0.396 &  31.13 / 0.8515  / 0.206 \\ 
         &  ESRGAN \cite{Wang_2018_ECCV_Workshops} & 31.20M & 26.63 / 0.7720 / 0.112 & 25.50 / 0.6315 / \textcolor[rgb]{0,0,1}{0.173} &  28.26 / 0.7746  / 0.105 \\ 
         &  SPSR \cite{Ma_2020_CVPR} & 39.30M & 27.16 / 0.7736 / 0.104 & 25.72 / 0.6406 / \textcolor[rgb]{1,0,0}{0.168} &  28.37 / 0.7739  / 0.102 \\ 
        \hline
        \hline
        \multirow{4}*{VSR} &  RBPN \cite{Haris_2019_CVPR} & 12.77M & 29.75 / 0.8531 / 0.236 & 27.79 / 0.7179 / 0.408 &  31.27 / 0.8559  / 0.218 \\ 
         &  RSDN \cite{10.1007/978-3-030-58610-2_38} & 6.19M & 29.00 / 0.8347 / 0.240  & 26.38 / 0.6412 / 0.480 &  30.26 / 0.8365  / 0.223 \\ 
         &  MuCAN \cite{10.1007/978-3-030-58607-2_20} & 25.66M & 29.90 / 0.8573 / 0.215 & 27.72 / 0.7162 / 0.400 &  31.24 / 0.8585  / 0.196 \\ 
         &  IconVSR \cite{Chan_2021_CVPR} & 8.71M & 29.75 / 0.8547 / 0.227 & 27.53 / 0.7089 / 0.408 &  31.41 / 0.8633  / 0.205 \\ 
        \hline
        \hline
         \multirow{7}*{Ref-SR} & TTSR \cite{Yang_2020_CVPR} & 145.66M & 27.53 / 0.7858 / 0.124 & 26.09 / 0.6567 / 0.201 & 28.80 / 0.7914 / 0.121 \\ 
        & TTSR-rec \cite{Yang_2020_CVPR} & 6.73M & 29.46 / 0.8471 / 0.231 & 27.68 / 0.7193 / 0.396 & 30.79 / 0.8485 / 0.216 \\ 
        & MASA \cite{Lu_2021_CVPR} & 18.99M & 26.66 / 0.7545 / 0.134 & 25.84 / 0.6468 / 0.220 & 28.05 / 0.7699 / 0.131 \\ 
        & MASA-rec \cite{Lu_2021_CVPR} & \textcolor[rgb]{1,0,0}{4.03M} & 29.67 / 0.8538 / 0.218 & 27.82 / 0.7335 / 0.375 & 31.12 / 0.8613 / 0.196 \\ 
         & $C^2$-Matching \cite{Jiang_2021_CVPR} & 14.11M & 29.29 / 0.8384 / 0.158 & 27.24 / 0.7039 / 0.281 &  30.57 / 0.8396   / 0.153  \\ 
        & $C^2$-Matching-rec \cite{Jiang_2021_CVPR} & 8.87M & 29.89 / 0.8609 / 0.212 & 27.99 / 0.7377 / 0.370 &  31.21 / 0.8628  / 0.194  \\ 
        \hline
        \hline
         &  MVSRnet & 19.42M & 28.83 / 0.8372 / \textcolor[rgb]{1,0,0}{0.085}  & 26.90 / 0.7080 / 0.182 & 30.27 / 0.8462 / \textcolor[rgb]{1,0,0}{0.085} \\ 
         &  MVSRnet-\emph{light} & 9.67M  & 28.48 / 0.8260 / \textcolor[rgb]{0,0,1}{0.090}  & 26.83 / 0.7003 / 0.190 & 29.89 / 0.8337 / \textcolor[rgb]{0,0,1}{0.094} \\ 
         &  MVSRnet-rec & 14.18M & \textcolor[rgb]{1,0,0}{30.26} / \textcolor[rgb]{1,0,0}{0.8748} / 0.191  & \textcolor[rgb]{1,0,0}{28.03} / \textcolor[rgb]{1,0,0}{ 0.7457} / 0.360  &  \textcolor[rgb]{1,0,0}{31.60} / \textcolor[rgb]{1,0,0}{0.8775} / 0.179 \\ 
         &  MVSRnet-\emph{light}-rec & \textcolor[rgb]{0,0,1}{4.39M} &  \textcolor[rgb]{0,0,1}{30.12} / \textcolor[rgb]{0,0,1}{0.8712} / 0.199 & \textcolor[rgb]{0,0,1}{28.00} / \textcolor[rgb]{0,0,1}{0.7426} / 0.368  & \textcolor[rgb]{0,0,1}{31.45} / \textcolor[rgb]{0,0,1}{0.8735} / 0.186 \\ 
         
        \hline
        \end{tabular}
        }

        \label{tab:model_size}
    \end{table*}

\section{Compare to Multi-view Super-Resolution Method}
\label{compare_mvsr}
As far as we know, our MVSRnet is the first to solve the multi-view image super-resolution problem using low-resolution images. Existing multi-view SR related methods rarely consider this problem. Some methods focus on multi-view texture \cite{8885943} or light field images \cite{Zhang_2019_CVPR_2} and some other methods \cite{9286862, 10.1145/2964284.2967260} require high-resolution multi-view as a reference, which is not easily available in practice due to high storage costs and bandwidth constraints. The most similar method to ours is SASRnet \cite{DBLP:conf/mm/SunC0Z21}. It uses two adjacent low-resolution views to generate high-resolution source views and a novel view. In Tanks and Temples and BlendedMVS dataset, they have a source view PSNR of 29.29dB and 27.27dB, respectively, which is much lower than our 30.26dB and 28.03dB as shown in Table~\ref{tab:compare_mvsr}. Moreover, SASRnet fails in GTAV and generate dark images because their depth prediction module is sensitive to depth scale. 

\section{Network Structures}
\label{network_structures}

We illustrate the network structure of the Dynamic High-Frequency Search network in Table~\ref{tab:dhfs}. It first uses a feature extractor to extract the features of multi-view reference images and near-view reference images at 1$\times$, 2$\times$, and 4$\times$ scales, respectively. 
Then three Raw Selection Modules (RSM) and three Adaptation Selection Modules (ASM) are used to select the useful features from the reference features of different scales. We adopt the same structure as \cite{Jiang_2021_CVPR} for the discriminator.

\section{Model Size}
\label{model_size}
This section compares the number of training parameters in Table~\ref{tab:model_size}. We provide two models, MVSRnet-\emph{light} and MVSRnet-\emph{light}-rec, for comparing model parameters. Specifically, for MVSRnet-\emph{light}, we reduce the number of channels in the feature extractor and residual layer to half and add three more residual layers to feature extractor.
We also include the parameter of the discriminator in the total parameters for the GAN-based model.
As shown in Table~\ref{tab:model_size}, the parameters of MVSRnet-\emph{light}-rec are only 0.36M more than MASA-rec, but our PSNR is 0.45dB, 0.18dB, and 0.33dB more than MASA-rec in Tanks and Temple, BlendedMVS, and GTAV, respectively.
In summary, MVSRnet-\emph{light} not only has fewer parameters, but achieves better results than the state of the arts to a large extent.

\section{More Visual Comparison}
\label{visual_comparison}
\begin{figure*}[t]
    \centering
    \includegraphics[width=0.88\linewidth]{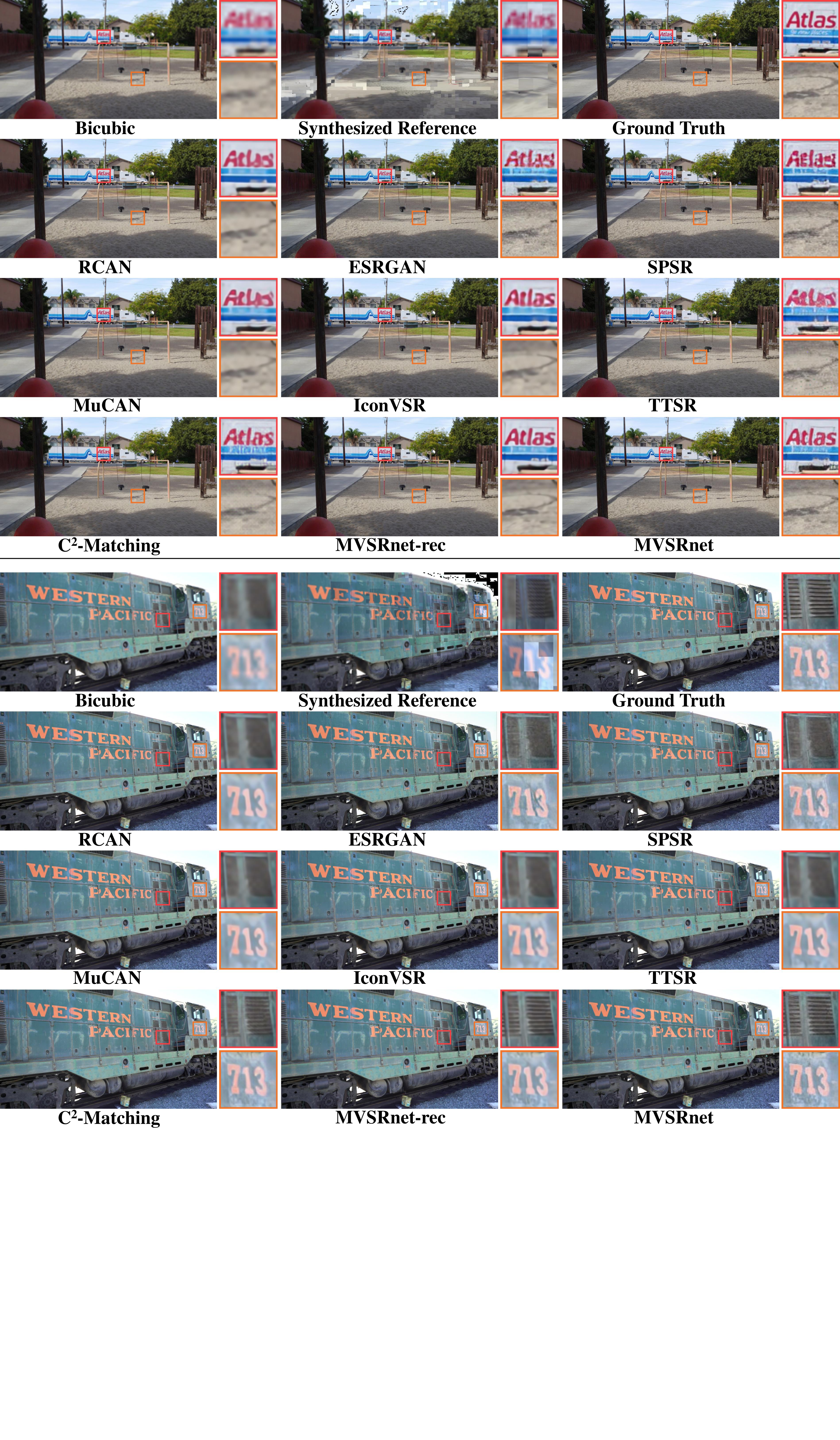}
    \caption{
    More qualitative comparisons with soft-of-the-art SR methods on the Tanks and Temple \cite{DBLP:journals/tog/KnapitschPZK17} dataset.
    }
    \label{009_supplementary_000}
\end{figure*}

\begin{figure*}[t]
    \centering
    \includegraphics[width=0.88\linewidth]{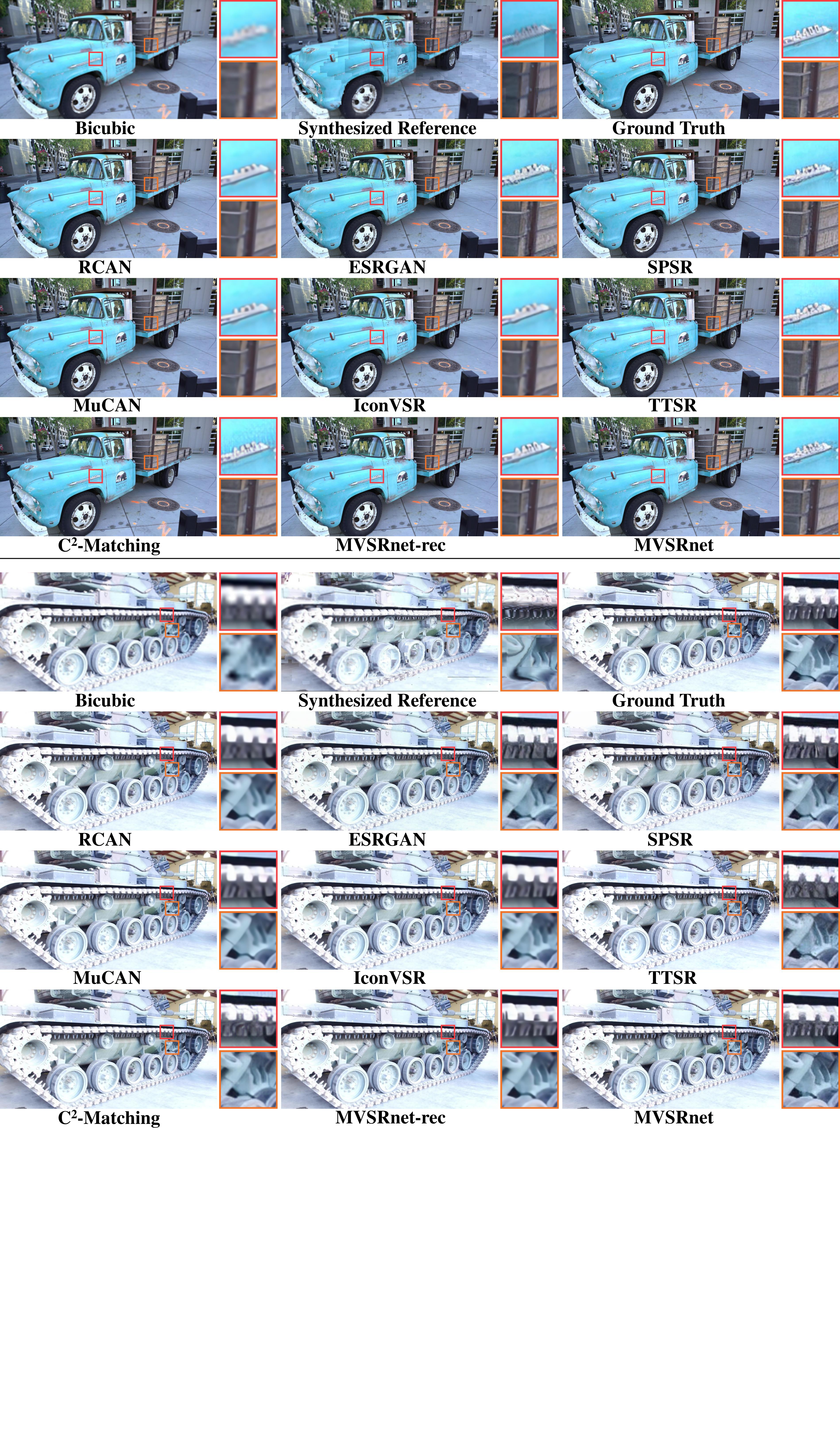}
    \caption{
    More qualitative comparisons with soft-of-the-art SR methods on the Tanks and Temple  \cite{DBLP:journals/tog/KnapitschPZK17} dataset.
    }
    \label{009_supplementary_001}
\end{figure*}

\begin{figure*}[t]
    \centering
    \includegraphics[width=0.88\linewidth]{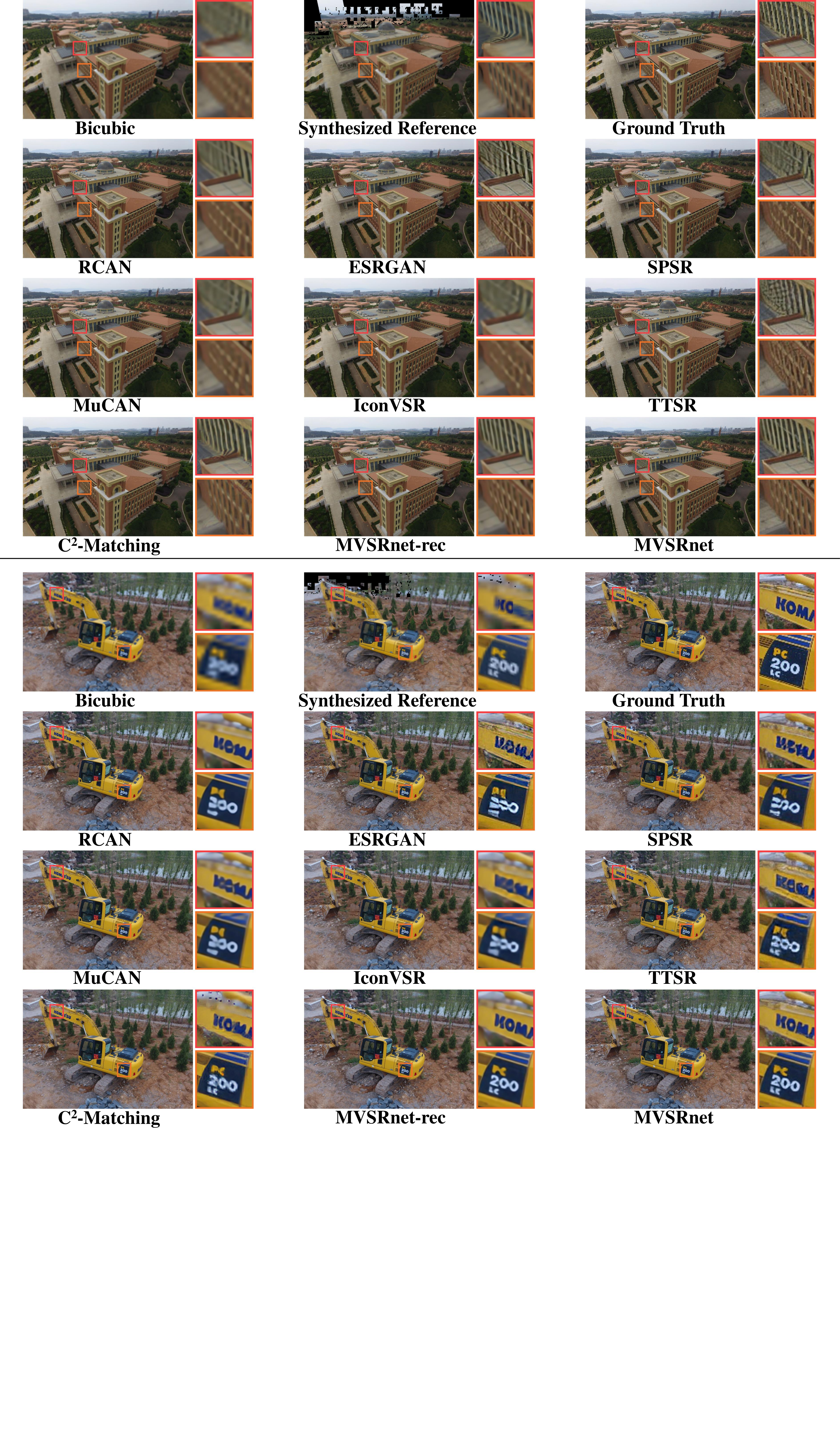}
    \caption{
    More qualitative comparisons with soft-of-the-art SR methods on the BlendedMVS  \cite{Yao_2020_CVPR} dataset.
    }
    \label{009_supplementary_002}
\end{figure*}

\begin{figure*}[t]
    \centering
    \includegraphics[width=0.88\linewidth]{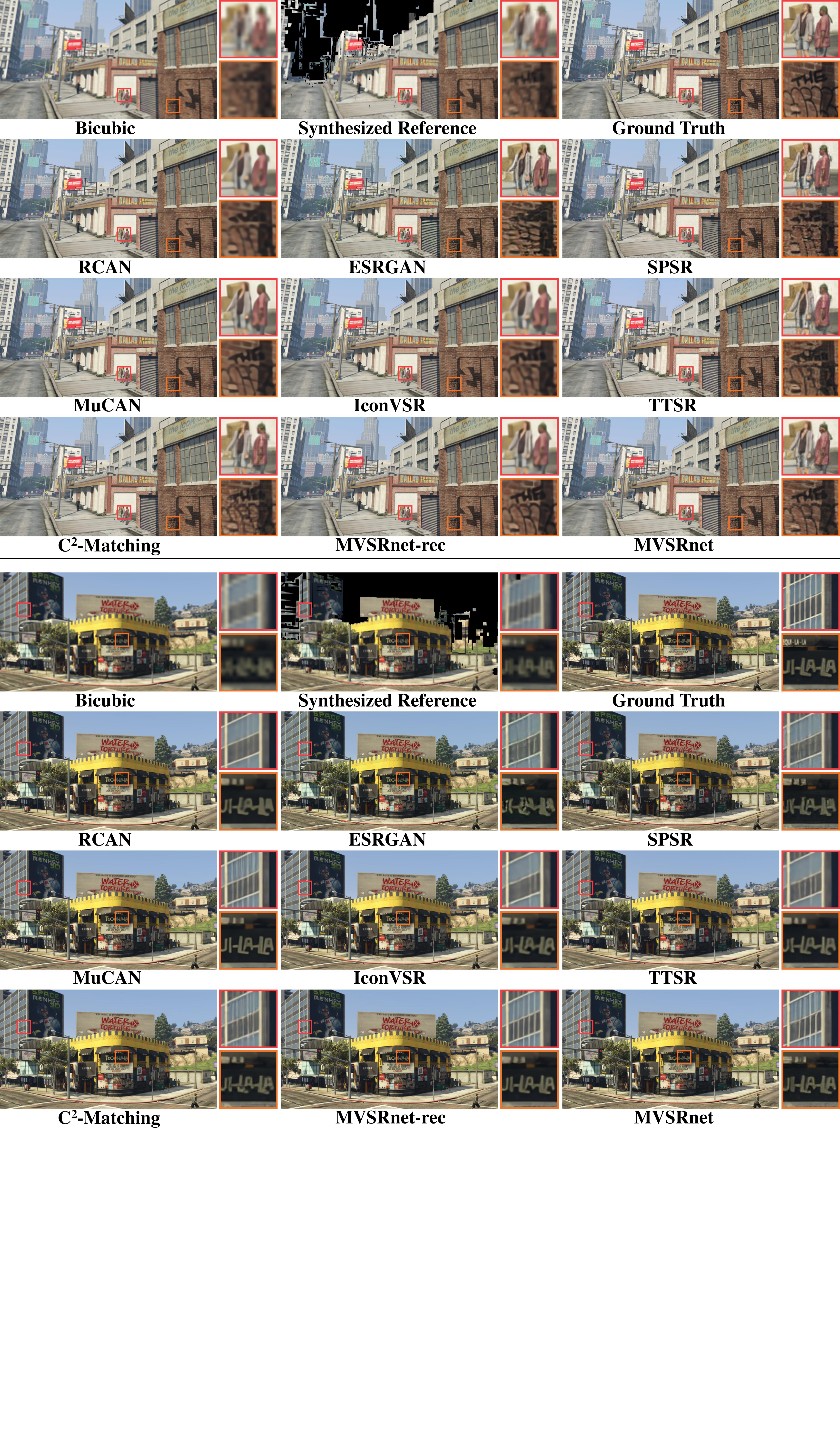}
    \caption{
    More qualitative comparisons with soft-of-the-art SR methods on the GTAV  \cite{Huang_2018_CVPR} dataset.
    }
    \label{009_supplementary_003}
\end{figure*}

In this section, we provide more visual comparisons with SISR methods (RCAN \cite{zhang2018rcan}, ESRGAN \cite{Wang_2018_ECCV_Workshops}, and SPSR \cite{Ma_2020_CVPR}), VSR methods (MuCAN \cite{10.1007/978-3-030-58607-2_20} and IconVSR \cite{Chan_2021_CVPR}), and Ref-SR methods (TTSR \cite{Yang_2020_CVPR} and $C^2$-Matching \cite{Jiang_2021_CVPR}).
Specifically, RCAN, MuCAN, IconVSR, and our MVSRnet-rec only use the reconstruction loss, and ESRGAN, SPSR, TTSR, $C^2$-Matching, and our MVSRnet are all GAN-based models.
The visual comparison of Tanks and Temple, BlendedMVS, and GTAV are shown in Figure~\ref{009_supplementary_000} and Figure~\ref{009_supplementary_001}, Figure~\ref{009_supplementary_002}, and Figure~\ref{009_supplementary_003}, respectively.

\section{Limitation}

We have two major limitations.
First, we assume static sceneries like in most multi-view work \cite{Riegler_2021_CVPR, Riegler2020FVS}, and the denser the input views, the better the scene geometry estimation. 
As a result, we will concentrate on the MVISR task of dynamic scenes and design a strategy for dealing with situations in which there are just a few multi-view images accessible. Second, we preprocess all LR multi-view images to create synthesized reference images, so an elaborated online reference selection approach should be proposed.

\clearpage

\end{document}